\documentclass[10pt,twocolumn]{article}
\usepackage{graphicx}
\usepackage{mathtools,amssymb,times,cite}
\usepackage{booktabs}
\usepackage{xcolor}
\newcommand{\op}[1]{\operatorname{#1}}
\newcommand{\bm}[1]{\mathbf{#1}} %Bold vectors and matrices

\newcommand\T{{\mathpalette\raiseT\intercal}}
\newcommand\raiseT[2]{%
\setbox0\hbox{$#1{#2}$}\raise\dp0\box0}

\usepackage{amsthm}
\newtheorem{thm}{Theorem}
\newtheorem{lem}[thm]{Lemma}

\theoremstyle{definition}

\title{\LARGE\textbf{Iterative Graph Filtering Network for 3D Human Pose Estimation}}
\author{Zaedul Islam and A. Ben Hamza\\
Concordia Institute for Information Systems Engineering\\
Concordia University, Montreal, QC, Canada
}
\date{}

\begin{document}
\maketitle

\begin{abstract}
Graph convolutional networks (GCNs) have proven to be an effective approach for 3D human pose estimation. By naturally modeling the skeleton structure of the human body as a graph, GCNs are able to capture the spatial relationships between joints and learn an efficient representation of the underlying pose. However, most GCN-based methods use a shared weight matrix, making it challenging to accurately capture the different and complex relationships between joints. In this paper, we introduce an iterative graph filtering framework for 3D human pose estimation, which aims to predict the 3D joint positions given a set of 2D joint locations in images. Our approach builds upon the idea of iteratively solving graph filtering with Laplacian regularization via the Gauss-Seidel iterative method. Motivated by this iterative solution, we design a Gauss-Seidel network (GS-Net) architecture, which makes use of weight and adjacency modulation, skip connection, and a pure convolutional block with layer normalization. Adjacency modulation facilitates the learning of edges that go beyond the inherent connections of body joints, resulting in an adjusted graph structure that reflects the human skeleton, while skip connections help maintain crucial information from the input layer's initial features as the network depth increases. We evaluate our proposed model on two standard benchmark datasets, and compare it with a comprehensive set of strong baseline methods for 3D human pose estimation. Our experimental results demonstrate that our approach outperforms the baseline methods on both datasets, achieving state-of-the-art performance. Furthermore, we conduct ablation studies to analyze the contributions of different components of our model architecture and show that the skip connection and adjacency modulation help improve the model performance.
\end{abstract}


\section{Introduction}
The goal of 3D human pose estimation is to predict the 3D locations of human body joints in the camera coordinate system from images or videos, with the aim of providing a way to interpret human movements and actions in computer vision applications, such as action recognition~\cite{Song2021Survey}, human-computer interaction, sports performance analysis, and pedestrian behavior analysis~\cite{zhao2019accurate}. Pose estimation can also be used in various healthcare applications, such as monitoring the physical therapy progress of patients or detecting abnormalities in movement patterns, and assisted living in retirement homes~\cite{rougier2006monocular}.

Despite significant progress in recent years~\cite{Liu2022Survey}, 3D human pose estimation remains a challenging task. This is largely attributed to two main challenges: (i) self-occlusions that occur when a body part is obscured by another part, and hence making it difficult for the model to accurately estimate the position of the occluded body part; and (ii) depth ambiguity that arises due to occlusions, self-occlusions, and variations in body shape, where there can be multiple 3D poses that correspond to the same 2D projection of a person in an image.

Recent approaches to 3D human pose estimation have focused on improving the accuracy and robustness of existing methods, as well as addressing the challenges of occlusion and depth ambiguity. These approaches can be grouped into two main categories: one-stage and two-stage methods. One-stage methods~\cite{zhou2016deep, park20163d, sun2018integral, pavlakos2017coarse, sun2017compositional, yang20183d, chen2020towards, lee2018propagating}, also known as direct regression methods, aim to directly predict the 3D joint locations from an input image or video without requiring any intermediate predictions. However, these methods usually suffer from depth ambiguity, which arises because the 3D pose estimation problem is inherently under-constrained, meaning that there are multiple possible 3D poses that can explain the same 2D observations. Also, they do not perform well when dealing with complex poses or occlusions. On the other hand, two-stage methods~\cite{martinez2017simple,cai2019exploiting,ci2019optimizing, pavllo20193d,wu20203d,xu2020deep,choi2020pose2mesh,wang2020motion,liu2020learning,liu2020attention,zhao2019semantic,zou2020high,liu2020comprehensive,quan2021higher,zou2021modulated,zou2021compositional,lee2022multi, zhang2022group}, also known as indirect regression methods, first predict intermediate representations such as 2D joint locations and then use them to predict the 3D joint locations. These methods are usually more accurate than one-stage methods, particularly when combined with robust 2D joint detectors, since they can better handle the depth ambiguity and occlusions.

In traditional 3D human pose estimation methods~\cite{martinez2017simple}, each joint in the human skeleton is localized independently, and the spatial relationship between the joints is not considered. This can lead to incorrect poses, especially in cases where joints occlude each other. To tackle this potential weakness, graph convolutional networks (GCNs) have recently shown promising results in 3D human pose estimation~\cite{zhao2019semantic}, yielding improved performance over traditional methods. By naturally modeling the skeleton structure of the human body as a graph, where each joint is represented as a node and the edges represent the relationships between the joints, GCNs are able to capture the spatial relationships between joints and learn an efficient representation of the underlying pose. One of the main benefits of using GCNs is that they can capture the dependencies between the joints. The position of each joint is dependent on the position of other joints in the body, and GCNs can model these dependencies explicitly. This allows the model to reason about the body as a connected system and to make accurate predictions even when some joints are occluded. Despite their promising results, GCNs suffer from several shortcomings~\cite{liu2020comprehensive}. First, they use a shared weight matrix (i.e., the same set of weights is used for all nodes in the graph) to determine the importance of neighboring joints when computing the representation of a particular joint. However, the shared weight matrix can be limited in its ability to capture complex relationships between joints in a human body, and may not be able to capture the nuances of joint relationships. This can limit the expressiveness of the model, making it difficult to learn more complex relationships between joints, thereby resulting in suboptimal performance~\cite{zou2021modulated}. Second, GCNs without skip connections are limited by the receptive field of their filters, which only allow them to capture information from the local neighborhood of each joint in the graph. Skip connections allow for the flow of information to bypass certain layers in the network and improve the modeling of long-range dependencies in the graph without sacrificing performance. Third, GCNs that use only local information can lead to over-smoothing~\cite{chen2020simple, xu2018representation, klicpera2018predict, chen2022bag}, where the learned node representations become too similar, especially when the network depth increases.

More recently, Transformer-based models~\cite{Lin2021Metro,Zhao2022graformer,Zheng2021PoseFormer} have been shown to be an effective approach for 3D human pose estimation due to their ability to capture long-range dependencies between the joints. The basic idea is to treat the 3D poses of human bodies as a sequence of tokens, each of which represents a joint in the human body, and the goal is to predict the 3D coordinates of each joint. However, there are some challenges associated with using Transformers for 3D human pose estimation, such as the need for large amounts of training data and the computational cost of processing long sequences of joint tokens. To mitigate some of these challenges, Zhuang \textit{et al.}~\cite{Liu2022convnext} introduce ConvNeXt, an architecture that solely utilizes CNN modules and performs competitively with Transformers in both accuracy and scalability, yielding superior performance over the Swin Transformer~\cite{liu2021swin}. Unlike the Swin Transformer, specialized modules like shifted window attention or relative position biases are not needed for ConvNeXt. The ConvNeXt residual block is fully convolutional, and uses a layer normalization, followed by the Gaussian Error Linear Unit activation function. This motivates us to explore and eventually design a variant of the ConvNeXt block for skeleton-based representation learning.

In this paper, we propose a Gauss-Seidel graph neural network (GS-Net), an architecture that incorporates a skip connection, weight and adjacency modulation, and a variant of the ConvNeXt residual block while upholding the flexibility of models intended for broad applications, including 3D human pose estimation, which is the primary focus of this work. Using an iterative graph filtering approach for solving a linear system of sparse equations, we start by deriving the layer-wise propagation rule of the proposed network via the Gauss-Seidel iterative method. Our framework follows the two-stage paradigm for 3D human pose estimation, where GS-Net is employed as a lifting network to predict the 3D pose locations from 2D predictions. Inspired by the regularized elastic net regression method, we train our model using a flexible loss function defined as a weighted sum of the mean squared and mean absolute errors between the 3D ground truth coordinates and estimated 3D joint coordinates over a training set consisting of human poses. Experiments on two large-scale datasets verify the effectiveness of our model using both quantitative and qualitative evaluations. Moreover, our ablation studies highlight the potential for further improvements through the incorporation of skip connections and adjacency modulation. Our contributions can be summarized as follows:
\begin{itemize}
\item We propose a novel Gauss-Seidel graph neural network (GS-Net) whose layer-wise propagation rule is obtained by iteratively solving graph filtering with Laplacian regularization via the Gauss-Seidel iterative method.
\item We design a network architecture comprised of weight and adjacency modulation, skip connection, and a variant of the ConvNeXt residual block.
\item We demonstrate through extensive experiments and ablation studies the effectiveness and generalization ability of the proposed GS-Net model, achieving competitive performance on two benchmark datasets.
\end{itemize}

\smallskip\noindent The remainder of this paper is organized as follows. In Section 2, the related work is discussed, highlighting the key findings and limitations of prior work. Section 3 presents the methodology, including the problem formulation, propagation rule, model architecture, and model training and prediction. The experimental setup and results are presented in Section 4. Finally, we conclude in Section 5 with a summary of our main contributions and their implications for future work directions.

\section{Related Work}
The basic goal of 3D human pose estimation is to determine the 3D positions of the joints of a person in a given image or video. These joints can include the head, torso, arms, legs and hands, and the accurate estimation of 3D human pose can provide valuable insights into human movement.

\medskip\noindent\textbf{3D Human Pose Estimation.}\quad As mentioned earlier, there are two general approaches for 3D human pose estimation: one-stage and two-stage. In the one-stage approach~\cite{pavlakos2017coarse,park20163d,chen20173d,tome2017lifting,tekin2017learning,chen2020towards}, 3D human pose estimation is performed in a single step without any intermediate processing. This approach typically involves training a deep learning model such as a convolutional neural network or a graph neural network to directly predict the 3D pose of the person in the image or video. The network takes the input image or video frame as input and outputs the 3D pose as a set of joint coordinates. The inherent ambiguity of 2D-to-3D mapping can, however, lead to multiple possible 3D pose solutions for a given 2D input. For example, Toshev \textit{et al.}~\cite{Toshev2014DeepPose} introduce the application of Deep Neural Networks (DNNs) to human pose estimation, leveraging a DNN-based regression approach to joint coordinates and a cascaded regressor system that provides the advantage of holistic pose reasoning and context awareness. Pavlakos \textit{et al.}~\cite{pavlakos2017coarse} uses a coarse-to-fine method to predict the 3D pose, where the coarse predictions guide the fine-grained predictions. This method works by first generating a coarse 3D volumetric representation of the human body using a 3D convolutional neural network (CNN). The coarse volumetric representation is then used to guide the fine-grained prediction of the 3D pose by training a 2D CNN to regress the 2D joint locations, followed by a 3D refinement network that refines the predicted 3D pose by taking into account the coarse volumetric representation. Another common approach to 3D human pose estimation is multi-view fusion, which leverages multiple camera views and fuses 2D poses to generate 3D pose estimates. Qiu \textit{et al.}~\cite{Qiu2019CrossView} present an novel approach for estimating 3D human poses from multiple calibrated cameras, utilizing a CNN-based multi-view feature fusion method to enhance 2D pose estimation accuracy and a recursive pictorial structure model to estimate 3D poses from the multi-view 2D poses. He \textit{et al.}~\cite{He2020Epipolar} introduce the epipolar transformer, enabling 2D pose detectors to utilize 3D-aware features by fusing features along epipolar lines of neighboring views, along with a recursive pictorial structure model to reconstruct the 3D pose from multi-view 2D poses. Liu \textit{et al.}~\cite{Liu2021DeepDual} present a dual consecutive network for multi-frame person pose estimation, leveraging a pose temporal merger and a pose residual fusion module to extract contextual information from adjacent frames, resulting in localized search ranges and improved accuracy in locating keypoints. Also, Liu \textit{et al.}~\cite{Liu2022FAMI} explore the multi-frame human pose estimation task by emphasizing the effective utilization of temporal contexts through feature alignment and complementary information mining, achieved by introducing a hierarchical coarse-to-fine network that progressively aligns supporting frame features with the key frame feature. While 3D pose estimation typically focuses on predicting the positions of skeleton joints, the task of 3D human body mesh recovery~\cite{Sun2019DSD} from monocular images aims to reconstruct the complete mesh representation of the human body, encompassing both pose and shape.

In the two-stage approach~\cite{martinez2017simple, cai2019exploiting, ci2019optimizing, pavllo20193d, wu20203d, xu2020deep, choi2020pose2mesh, zou2020high, wang2020motion, liu2020learning, liu2020attention,zhao2019semantic,Li2022MHFormer}, 3D human pose estimation is performed in two steps: first, the 2D keypoints of the person are detected in the input image or video frame using an off-the-shelf 2D pose detector~\cite{chen2018cascaded,sun2019deep}, and then the 3D pose is estimated from the 2D keypoints using a 2D-to-3D lifting network. For instance, Martinez \textit{et al.}~\cite{martinez2017simple} propose a simple multilayer neural network architecture with a building block comprised primarily of a linear layer, followed by batch normalization, dropout, a Rectified Linear Unit (RELU) activation function, as well as a residual connection to help improve generalization performance and reduce training time. Li \textit{et al.}~\cite{Li2022MHFormer} introduce multi-hypothesis Transformer, a novel three-stage framework based on the Transformer architecture, designed to address the ambiguous inverse problem of 3D human pose estimation from monocular videos by generating multiple pose hypotheses in the spatial domain and facilitating communication between them in both independent and mutual manners in the temporal domain. Our approach falls under the category of two-stage methods, and we employ the proposed GS-Net model as a lifting network. We first estimate the 2D joint coordinates in the input image using a 2D pose estimator, and then we use these 2D joint locations as input to GS-Net to predict 3D joint coordinates. A key advantage of using GS-Net as our lifting network is that we can leverage the key components (i.e., skip connection, weight and adjacency modulation, and ConvNeXt) present in the model architecture to effectively estimate the 3D joint positions.

\medskip\noindent\textbf{Graph Convolutional Network based Methods.}\quad The basic idea behind GCN-based methods for 3D human pose estimation is to represent the human body a graph, where each node represents a joint and the edges represent the skeletal connections between them~\cite{zhao2019semantic,zou2020high,liu2020comprehensive,quan2021higher,zou2021modulated,zou2021compositional,lee2022multi,zhang2022group}. Zhao \textit{et al.}~\cite{zhao2019semantic} propose SemGCN, a semantic graph convolutional network that has the ability to learn and encode semantic information such as both local and global node relationships, which may not be explicitly represented in the graph structure. Zou \textit{et al.}~\cite{zou2020high} design a high-order graph convolutional network to capture high-order dependencies between body joints by considering neighbors that are multiple hops away when updating the joint features in an effort to reduce the uncertainty that arises from occlusion or depth ambiguity. In~\cite{zou2021modulated}, a modulated GCN comprised of weight modulation and adjacency modulation is proposed, where weight modulation enables the learning of unique modulation vectors for individual nodes and adjacency modulation modifies the graph structure to account for additional edges beyond the human skeleton that can be modeled. Building upon modulated GCN, Lee \textit{et al.}~\cite{lee2022multi} design multi-hop modulated GCN, an architecture in which the features of multi-hop neighbors are modulated and fused by a learnable matrix designed to assign higher weights to features of neighbors that are closer in hop distance. Zhang \textit{et al.}~\cite{zhang2022group} introduce GroupGCN, which is inspired by the concept of group convolution in convolutional neural networks. The GroupGCN architecture includes group convolution that is used to ensure that each group has its own weight matrix and spatial aggregation kernel, and group interaction that allows features to interact between groups and take into account global information for better performance.

Our approach differs from these GCN-based approaches in that instead of using GCN as a 2D-to-3D lifting network, we design a new graph neural network with skip connections, together with weight and adjacency modulation. We leverage skip connections, which are integrated by design in our network architecture, allowing the network to reuse lower-level features in higher-level layers, thereby helping the model to learn more complex representations. In addition, we employ a variant of the ConvNeXt block in our network architecture. ConvNeXt has demonstrated competitive accuracy and scalability~\cite{Liu2022convnext} compared to Transformers, while retaining the simplicity and efficiency of standard convolutional neural networks.

\section{Proposed Method}

\subsection{Preliminaries and Problem Statement}
\noindent\textbf{Basic Notions.}\quad Let $\mathcal{G}=(\mathcal{V},\mathcal{E}, \bm{X})$ be an attributed graph, where $\mathcal{V}=\{1,\ldots,N\}$ is a set of nodes that correspond to body joints, $\mathcal{E}$ is the set of edges representing connections between two neighboring body joints, and $\bm{X}=(\bm{x}_{1},...,\bm{x}_{N})^{\T}$ is an $N\times F$ feature matrix of node attributes whose $i$-th row $\bm{x}_{i}$ is an $F$-dimensional feature vector associated to node $i$. The graph structure is encoded by an $N\times N$ adjacency matrix $\bm{A}$ whose $(i,j)$-th entry is equal to 1 if there the edge between neighboring nodes $i$ and $j$, and 0 otherwise. We denote by $\hat{\bm{A}}=\bm{D}^{-1/2}\bm{A}\bm{D}^{-1/2}$ the normalized adjacency matrix, where $\bm{D}=\op{diag}(\bm{A}\bm{1})$ is the diagonal degree matrix and $\bm{1}$ is an $N$-dimensional vector of all ones.

\medskip\noindent\textbf{Gauss-Seidel Method.}\quad Let $\bm{\Phi}$ be an $N\times N$ matrix whose diagonal entries are all nonzero, and consider the matrix decomposition $\bm{\Phi}=\bm{\Lambda}-(\bm{E}+\bm{F})$, where $\bm{\Lambda}$ is the diagonal matrix of $\bm{\Phi}$, $\bm{E}$ is its strictly lower triangular part, and $\bm{F}$ is its strictly upper triangular part. Given a vector $\bm{x}\in\mathbb{R}^{N}$, the Gauss-Seidel iteration~\cite{Saad:03} for solving a matrix equation $\bm{\Phi}\bm{h}=\bm{x}$ is given by
\begin{equation}
\bm{h}^{(k+1)} = (\bm{\Lambda}-\bm{E})^{-1}\bm{F}\bm{h}^{(k)} + (\bm{\Lambda}-\bm{E})^{-1}\bm{x},
\end{equation}
where $\bm{h}^{(k)}$ and $\bm{h}^{(k+1)}$ are the $k$-th and $(k+1)$-th iterations of the unknown $\bm{h}$, respectively.

\medskip\noindent\textbf{Problem Formulation.}\quad Let $\mathcal{D}=\left\{\left(\mathbf{x}_{i}, \mathbf{y}_{i}\right)\right\}_{i=1}^{N}$ be a training set consisting of 2D joint positions $\bm{x}_{i}\in\mathcal{X}\subset\mathbb{R}^2$ and their associated ground-truth 3D joint positions $\bm{y}_{i}\in\mathcal{Y}\subset\mathbb{R}^3$. The aim of 3D human pose estimation is to learn the parameters $\bm{w}$ of a regression model $f: \mathcal{X} \rightarrow \mathcal{Y}$ by finding a minimizer of the following loss function
\begin{equation}	
\bm{w}^{*}=\arg\min_{\bm{w}}\frac{1}{N}\sum_{i=1}^{N}l(f(\bm{x}_{i}),\bm{y}_{i}),
\end{equation}
where $l(f(\bm{x}_{i}),\bm{y}_{i})$ is an empirical loss function defined by the learning task. Since human pose estimation is a regression task, we define $l(f(\bm{x}_{i}),\bm{y}_{i})$ as a weighted sum (convex combination) of the $\ell_2$ and $\ell_1$ loss functions
\begin{equation}	
l=(1-\alpha)\sum_{i=1}^{N}\Vert\bm{y}_{i}-f(\bm{x}_{i})\Vert_{2}^{2}+
\alpha\sum_{i=1}^{N}\Vert\bm{y}_{i}-f(\bm{x}_{i})\Vert_{1},
\label{eq:loss}
\end{equation}
where $\alpha\in [0,1]$ is a weighting factor that controls the mixing amount between the $\ell_2$ and $\ell_1$ penalties. It is worth pointing out that the proposed loss function draws inspiration from the penalty function used in the elastic net regression model~\cite{HuiZou:05}, which is a weighted combination of lasso and ridge regularization. When $\alpha=0$, the penalty function is equivalent to lasso regression, and when $\alpha=1$, it is equivalent to ridge regression. A key advantage of elastic net regression is that it reduces the impact of irrelevant predictors by shrinking their coefficients towards zero, unlike ridge regression which only reduces the size of the coefficients.

\subsection{Iterative Graph Filtering}
Graph filtering refers to the process of enhancing signals defined on graphs while preserving the underlying graph structure. It typically employs graph Laplacian regularization, which aims to incorporate the underlying graph structure into optimization problems to achieve desirable properties in the solutions while encouraging the filtered signal to be smooth and to vary smoothly across the graph. More specifically, the goal of graph filtering with Laplacian regularization is to minimize the following objective function
\begin{equation}
\mathcal{J}(\bm{H}) =\frac{1}{2}\Vert\bm{H}-\bm{X}\Vert_{F}^{2}+\frac{\beta}{2}\op{tr}(\bm{H}^{\T}\bm{L}\bm{H}),
\end{equation}
where $\bm{X}$ is the initial feature matrix, $\bm{H}$ is the filtered feature matrix, $\bm{L}=\bm{I}-\hat{\bm{A}}$ is the normalized Laplacian matrix, $\beta\in (0,1)$ is a regularization parameter, and $\Vert\cdot\Vert_{F}$ and $\op{tr}(\cdot)$ denote the Frobenius norm and trace operator, respectively. The first term on the right-hand side represents a fitting constraint or data fidelity, where a filtered feature matrix is expected to exhibit limited deviation from the initial feature matrix. The second term represents a smoothness constraint, which requires an effective filtered feature matrix to have consistent behavior in the vicinity of neighboring graph nodes (i.e., adjacent nodes should have similar node features). The trade-off parameter $\beta$ controls how much emphasis to put on data fidelity versus smoothness. A larger value of $\beta$ corresponds to a stronger regularization effect and a greater emphasis on smoothness of the filtered feature matrix at the expense of fitting the initial data matrix more closely, while a smaller value places more emphasis on the data fidelity term and results in a better fit to the initial feature matrix.

\noindent Taking the derivative of $\mathcal{J}$ and set it to zero, we have
\begin{equation}
\frac{\partial\mathcal{J}}{\partial\bm{H}}=\bm{H}-\bm{X}+\beta\bm{L}\bm{H}=\bm{0},
\end{equation}
which yields a system of sparse equations given by
\begin{equation}
(\bm{I}+\beta\bm{L})\bm{H}=\bm{X},
\end{equation}
where $\beta$ controls the smoothness of the filtered graph signal. It is important to point out that the matrix $\bm{I}+\beta\bm{L}$ is symmetric positive definite with minimal eigenvalue equal to 1 and maximal eigenvalue bounded from above by $1+2\beta$.

\medskip\noindent\textbf{Iterative Solution.}\quad Using the matrix decomposition for the Gauss-Seidel method, we can express $\bm{I}+\beta\bm{L}$ as the sum of a diagonal matrix, a strictly lower triangular matrix, and a strictly upper triangular matrix as follows:
\begin{equation}
\bm{I}+\beta\bm{L}=(1+\beta)\bm{I}-\beta\hat{\bm{A}}=\bm{\Lambda}_{\beta}-(\hat{\bm{A}}^{\T}_{\beta}+\hat{\bm{A}}_{\beta}),
\end{equation}
where $\bm{\Lambda}_{\beta}=(1+\beta)\bm{I}$ is a diagonal matrix (i.e., uniformly scaled identity matrix), and $\hat{\bm{A}}_{\beta}$ denotes the upper triangular matrix of $\beta\hat{\bm{A}}$. Since the uniformly scaled and normalized adjacency matrix $\beta\hat{\bm{A}}$ is symmetric with zero diagonal entries, its lower triangular part is $\hat{\bm{A}}^{\T}_{\beta}$.  Hence, the Gauss-Seidel iterative solution of $(\bm{I}+\beta\bm{L})\bm{H}=\bm{X}$ is given by
\begin{equation}
\bm{H}^{(k+1)}=(\bm{\Lambda}_{\beta}-\hat{\bm{A}}^{\T}_{\beta})^{-1}\hat{\bm{A}}_{\beta}\bm{H}^{(k)}+(\bm{\Lambda}_{\beta}-\hat{\bm{A}}^{\T}_{\beta})^{-1}\bm{X},
\label{Eq:Gauss-Seidel1}
\end{equation}
Since matrix inversion can be computationally expensive, especially for large matrices, a common approach to approximate the inverse of a matrix is to use a truncated series approximation such as the Neumann series expansion, which involves summing a finite number of terms of an infinite series, providing a good approximation of the inverse.
\begin{lem}
Let $\bm{S}$ be an $N\times N$ matrix with spectral radius $\rho(\bm{S})<1$. Then, $\bm{I}-\bm{S}$ is invertible and we have
\begin{equation}
(\bm{I}-\bm{S})^{-1}=\sum_{i=1}^{\infty} \bm{S}^{i}.
\end{equation}
\end{lem}
\smallskip\noindent Recall that the eigenvalues of a lower or upper triangular matrix are its diagonal entries. Hence, the spectral radius of the lower triangular matrix $\hat{\bm{A}}^{\T}_{\beta}-\beta\bm{I}$ is equal to $\beta$. Since $\beta\in (0,1)$, we can approximate the inverse of $\bm{\Lambda}_{\beta}-\hat{\bm{A}}^{\T}_{\beta}$ using Lemma 1 with first order approximation as follows:
\begin{equation}
(\bm{\Lambda}_{\beta}-\hat{\bm{A}}^{\T}_{\beta})^{-1}=(\bm{I}-(\hat{\bm{A}}^{\T}_{\beta}-\beta\bm{I}))^{-1}\approx (1-\beta)\bm{I} + \hat{\bm{A}}^{\T}_{\beta}.
\end{equation}
Therefore, the Gauss-Seidel iterative solution becomes
\begin{equation}
\begin{split}
\bm{H}^{(k+1)} &=((1-\beta)\bm{I} + \hat{\bm{A}}^{\T}_{\beta})\hat{\bm{A}}_{\beta}\bm{H}^{(k)}\\
 & \quad{}+((1-\beta)\bm{I}+\hat{\bm{A}}^{\T}_{\beta})\bm{X}.
\end{split}
\label{Eq:Gauss-Seidel}
\end{equation}

\subsection{Gauss-Seidel Network}
Propagation rules in graph neural networks such as GCNs generally define how information is passed between nodes in a graph during the forward pass of the network. These rules are used to update the feature representations of nodes in the graph based on information from their neighbors, and are a critical component in the learning process of graph neural networks. Motivated by the Gauss-Seidel iterative solution \eqref{Eq:Gauss-Seidel} of graph filtering with Laplacian regularization, we propose a Gauss-Seidel network (GS-Net) with the following layer-wise propagation rule:
\begin{equation}
\begin{split}
\bm{H}^{(\ell+1)} &= \sigma\left(((1-\beta)\bm{I} + \hat{\bm{A}}^{\T}_{\beta})\hat{\bm{A}}_{\beta}\bm{H}^{(\ell)}\bm{W}^{(\ell)} \right. \\
  &\qquad \left. {}+ ((1-\beta)\bm{I}+\hat{\bm{A}}^{\T}_{\beta})\bm{X}\widetilde{\bm{W}}^{(\ell)}\right),
\end{split}
\end{equation}
where $\bm{W}^{(\ell)}\in\mathbb{R}^{F_{\ell}\times F_{\ell+1}}$ and $\widetilde{\bm{W}}^{(\ell)}\in\mathbb{R}^{F\times F_{\ell+1}}$ are learnable weight matrices, $\sigma(\cdot)$ is an element-wise nonlinear activation function such as the Gaussian Error Linear Unit (GELU), $\bm{H}^{(\ell)}\in\mathbb{R}^{N\times F_{\ell}}$ is the input feature matrix of the $\ell$-th layer and $\bm{H}^{(\ell+1)}\in\mathbb{R}^{N\times F_{\ell+1}}$ is the output feature matrix. The input of the first layer is the initial feature matrix $\bm{H}^{(0)}=\bm{X}$.

Note that the second term on the right-hand side of the propagation rule is basically a skip connection that provides a effective way to pass information directly from the initial feature matrix to the next network layer, without any nonlinear transformation in between. This allows the information to be preserved and can help improve the flow of information through the network, enabling the model to learn richer, more expressive feature representations.

\medskip\noindent\textbf{Weight Modulation.}\quad One of the main shortcomings of GCNs is that they share the weight matrix, meaning that they treat all nodes in the graph equally, without considering their individual characteristics. This can lead to over-smoothing, where the learned representations of nodes become too similar, reducing the ability of the model to distinguish between different nodes in the graph. To circumvent this limitation, we leverage weight modulation~\cite{zou2021modulated}, which employs a common weight matrix, but a learnable weight modulation vector is introduced for each node, which scales the weight matrix according to the local topology of the node. This enables each node to have a unique and adaptive representation, which can capture more fine-grained information about the node's local structure and its position in the graph. Hence, the layer-wise propagation rule with weight modulation can be written as
\begin{equation}
\begin{split}
\bm{H}^{(\ell+1)} &= \sigma\left(((1-\beta)\bm{I} + \hat{\bm{A}}^{\T}_{\beta})\hat{\bm{A}}_{\beta}(\bm{M}^{(\ell)}{\odot}(\bm{H}^{(\ell)}\bm{W}^{(\ell)})) \right. \\
  &\qquad \left. {}+ ((1-\beta)\bm{I}+\hat{\bm{A}}^{\T}_{\beta})(\bm{M}^{(\ell)}{\odot}(\bm{X}\widetilde{\bm{W}}^{(\ell)}))\right),
\end{split}
\end{equation}
where $\bm{M}^{(\ell)}\in\mathbb{R}^{N\times F_{\ell+1}}$ is a learnable weight modulation matrix, and $\odot$ denotes element-wise multiplication.

\medskip\noindent\textbf{Adjacency Modulation.}\quad We modulate the normalized adjacency matrix to capture not just the interactions between adjacent nodes, but also the relationships between distant nodes beyond the natural connections of body joints~\cite{zou2021modulated}.

\begin{equation}
\check{\bm{A}}=\hat{\bm{A}}+\bm{Q},
\end{equation}
where $\bm{Q}^{(\ell)}\in\mathbb{R}^{N\times N}$ is a learnable adjacency modulation matrix. Since the skeleton graph is symmetric, we symmetrize the adjacency modulation matrix by taking the average of the matrix and its transpose, i.e., $(\bm{Q} + \bm{Q}^{T}) / 2$. Therefore, the layer-wise propagation rule of the Gauss-Seidel graph network with weight and adjacency modulation can be written as
\begin{equation}
\begin{split}
\bm{H}^{(\ell+1)} &= \sigma\left(((1-\beta)\bm{I} + \check{\bm{A}}^{\T}_{\beta})\check{\bm{A}}_{\beta}(\bm{M}^{(\ell)}{\odot}(\bm{H}^{(\ell)}\bm{W}^{(\ell)})) \right. \\
  &\qquad \left. {}+ ((1-\beta)\bm{I}+\check{\bm{A}}^{\T}_{\beta})(\bm{M}^{(\ell)}{\odot}(\bm{X}\widetilde{\bm{W}}^{(\ell)}))\right).
\end{split}
\end{equation}
In adjacency modulation, a weight modulation vector is introduced for each node, which is learned during training and is used to modulate the weights of the adjacency matrix. This enables the adjacency matrix to be dynamically adjusted based on the local node features, which can lead to better performance on 3D human pose estimation.

\begin{figure*}[!htb]
\centering
\includegraphics[scale=.27]{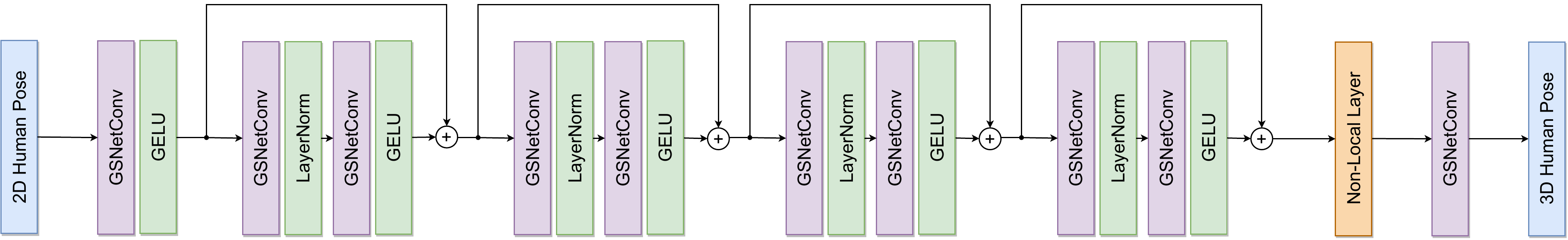}
\caption{Network architecture of the proposed GS-Net model for 3D human pose estimation. Our model accepts 2D pose coordinates (16 or 17 joints) as input and generates 3D pose predictions (16 or 17 joints) as output. We use ten Gauss-Seidel graph convolutional layers with four residual blocks. In each residual block, the first convolutional layer is followed by layer normalization, while the second convolutional layer is followed by a GELU activation function, except for the last convolutional layer, which is preceded by a non-local layer.}
\label{Fig:network_architecture}
\end{figure*}

\medskip\noindent\textbf{Model Architecture.}\quad Figure~\ref{Fig:network_architecture} illustrates the architecture of our proposed GS-Net model for 3D human pose estimation. The input to the model consists of 2D keypoints, obtained via an off-the-shelf 2D detector~\cite{chen2018cascaded}. Inspired by the architectural design of the ConvNeXt block~\cite{Liu2022convnext}, our residual block consists of two graph convolutional (GS-NetConv) layers. The first convolutional layer is followed by layer normalization, while the second convolutional layer is followed by a GELU activation function, which is a smoother version of ReLU and is commonly used in Transformers based approaches. This residual block is repeated four times. We also employ a non-local layer~\cite{wang2018non} before the last convolutional layer. The last layer of the network generates the 3D pose.

\medskip\noindent\textbf{Model Prediction.}\quad The output of the last graph convolutional layer of GS-Net contains the final output node embeddings, which are given by
\begin{equation}
\hat{\bm{Y}}=(\hat{\bm{y}}_{1},\dots,\hat{\bm{y}}_{N})^{\T}\in\mathbb{R}^{N\times 3},
\end{equation}
where $\hat{\bm{y}}_{i}$ is a 3D row vector of predicted 3D pose coordinates. This predicted set of 3D joint coordinates can be visualized in a 3D space, allowing for interactive manipulation and analysis of the pose.

\medskip\noindent\textbf{Model Training.}\quad The parameters (i.e., weight matrices for different layers) of the proposed GS-Net model for 3D human pose estimation are learned by minimizing the following loss function

\begin{equation}
\mathcal{L} =\frac{1}{N}\left[(1-\alpha)\sum_{i=1}^{N}\Vert\bm{y}_{i}-\hat{\bm{y}}_{i}\Vert_{2}^{2} + \alpha\sum_{i=1}^{N}\Vert\bm{y}_{i}-\hat{\bm{y}}_{i}\Vert_{1}\right],
\end{equation}
which is a weighted sum of the mean squared and mean absolute errors between the 3D ground truth coordinates $\bm{y}_{i}$ and estimated 3D
joint coordinates $\hat{\bm{y}}_{i}$ over a training set consisting of $N$ human poses. For the mean squared error, the squared differences between the predicted and ground truth coordinates are averaged, meaning that larger errors have a greater impact on the overall score. In other words, the mean squared error is more sensitive to outliers and penalizes larger errors more heavily than the mean absolute error, which is more robust to outliers and treats all errors equally. The weighting factor $\alpha$ balances the contribution of each loss term. When $\alpha=0$, our loss function reduces to the mean squared error (i.e., ridge regression) and when $\alpha=1$, it reduces to the mean absolute error (i.e., lasso regression).

\section{Experiments}
In this section, we present the results of our proposed framework against competing baselines for 3D human pose estimation. We begin by outlining the experimental setup, followed by providing an overview of evaluation protocols and implementation details. Then, we present both quantitative and qualitative results on two benchmark datasets using various evaluation metrics. We also conduct ablation studies on the significance of various components in our model in an effort to provide valuable insight into the effectiveness of the model. The code is available at: https://github.com/zaedulislam/GS-Net

\subsection{Experimental Setup}
\noindent\textbf{Datasets.}\quad We comprehensively evaluate our model on Human 3.6M~\cite{ionescu2013human3} and MPI-INF-3DHP~\cite{mehta2017monocular}, which are standard large-scale benchmark datasets for 3D human pose estimation.

\begin{figure}[!htb]
\centering
\footnotesize
\setlength{\tabcolsep}{1pt}
\begin{tabular}{ccccc}
\includegraphics[width=.65in]{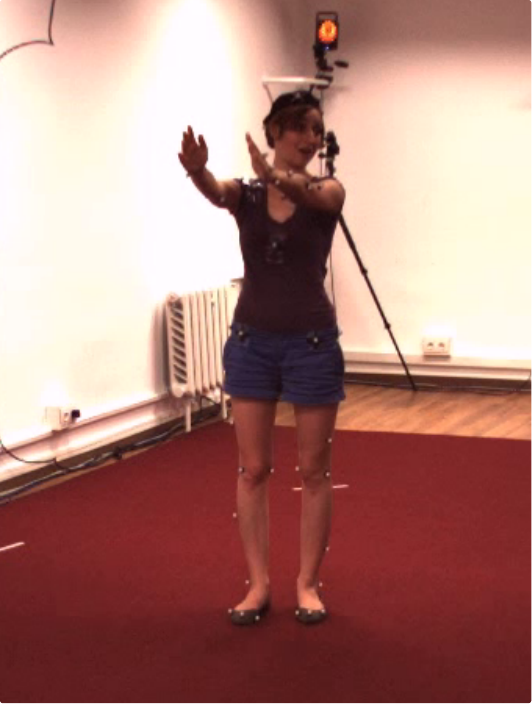} & \includegraphics[width=.65in]{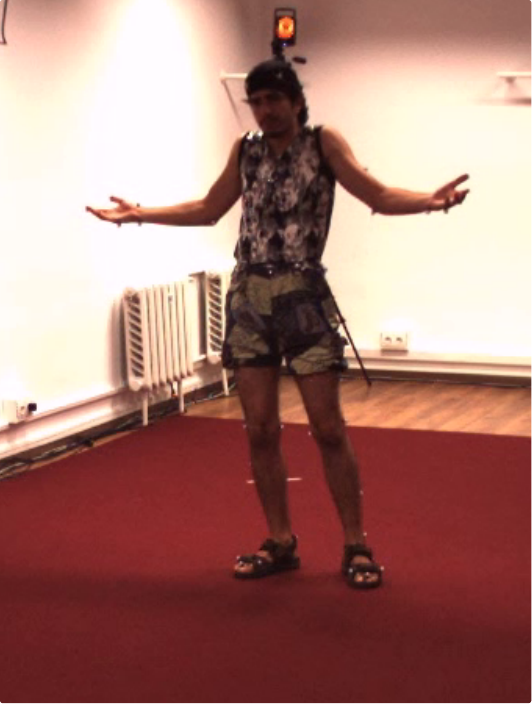} & \includegraphics[width=.65in]{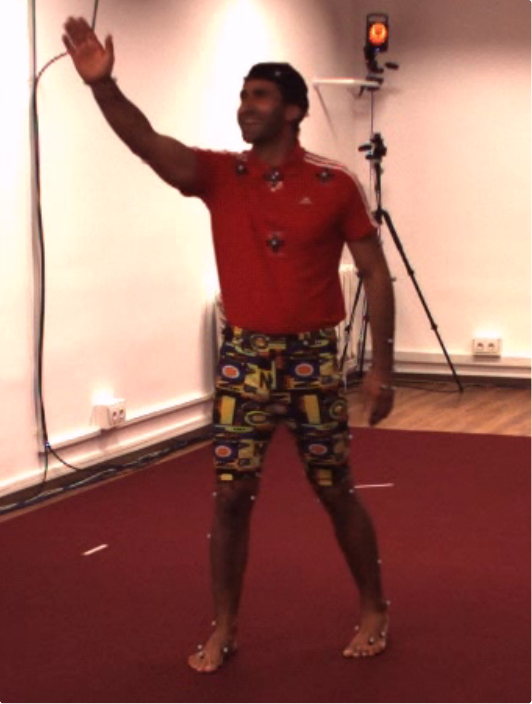} & \includegraphics[width=.65in]{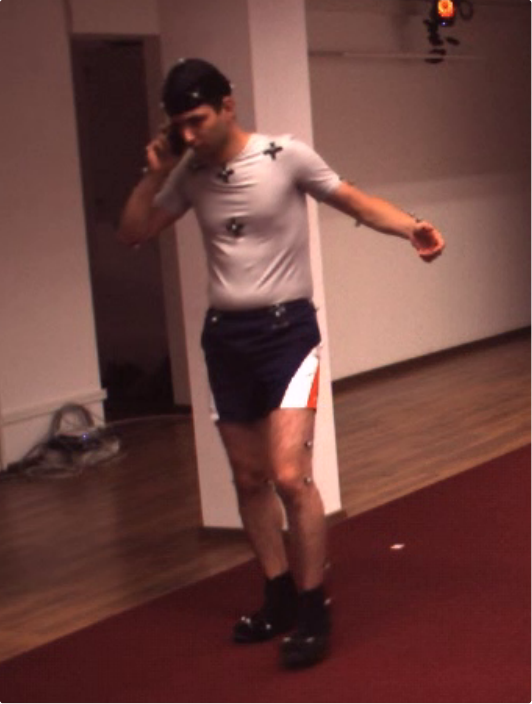} & \includegraphics[width=.65in]{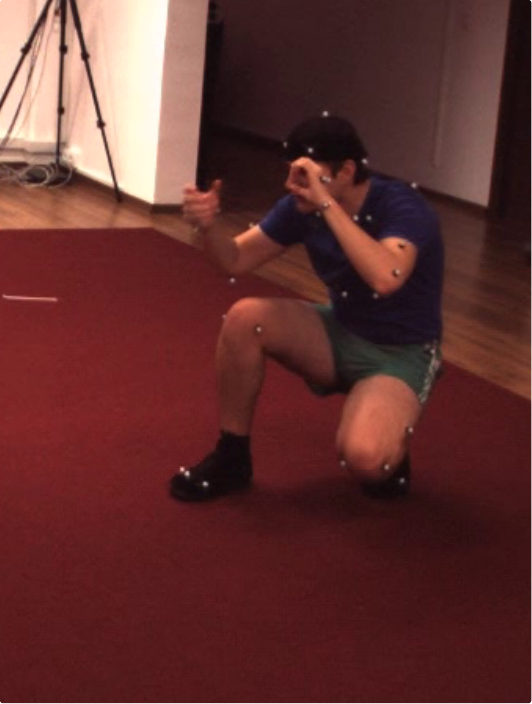} \\
Directions & Discussion & Greeting & Phone & Photo \\[1ex]
\includegraphics[width=.65in]{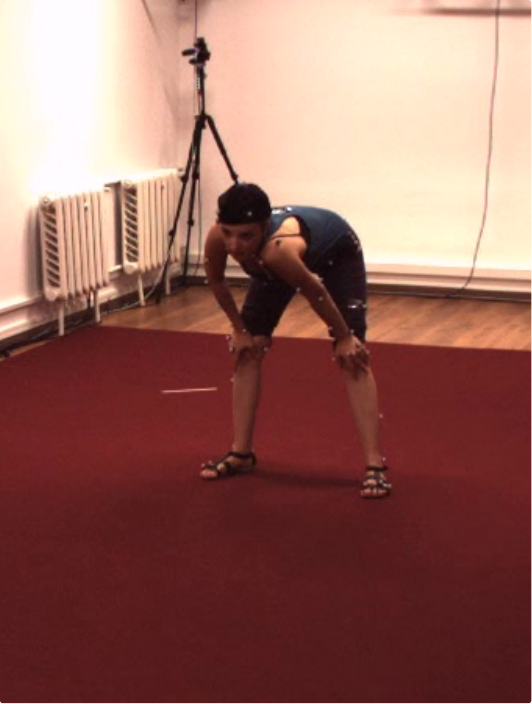} & \includegraphics[width=.65in]{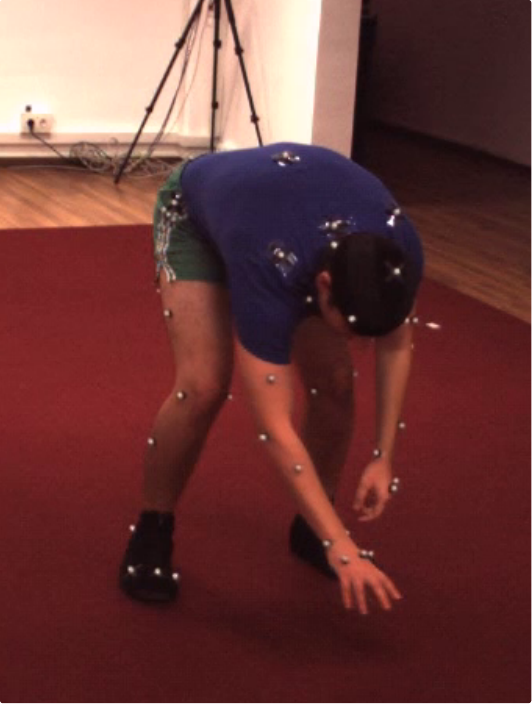} & \includegraphics[width=.65in]{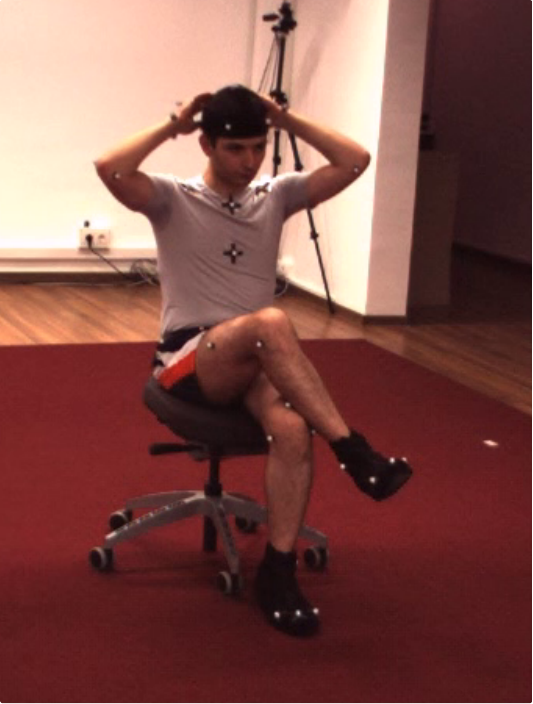} &
\includegraphics[width=.65in]{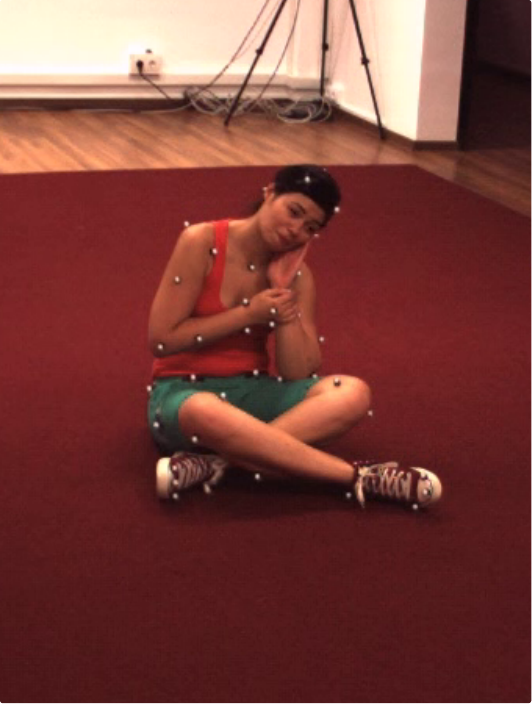} & \includegraphics[width=.65in]{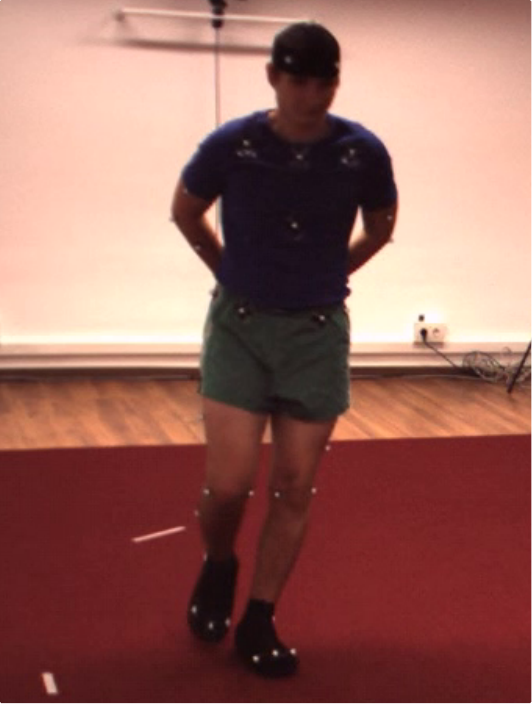} \\
Posing & Purchase & Sitting & Sitting Down & Walking\\[1ex]
\end{tabular}
\caption{Examples of actions performed by different actors in the Human3.6M dataset~\cite{ionescu2013human3}.}
\label{Fig:Actions}
\end{figure}

\smallskip
\noindent\textbf{\textsf{\footnotesize Human3.6M}} is a large-scale dataset comprised of 3.6 million images captured at 50Hz by 4 synchronized cameras in different positions and perspectives. A total of 11 professional actors (6 men and 5 women) perform 15 actions (Directions, Discussion, Eating, Greeting, Phoning, Posing, Purchases, Sitting, Sitting Down, Smoking, Photo, Waiting, Walk Dog, Walking, and Walk Together) in an interior setting, as depicted in Figure~\ref{Fig:Actions}. A motion capture system captures the annotations of precise 3D body joint coordinates, while projection with known intrinsic and extrinsic camera parameters yields the 2D poses. Annotated 3D joints are available for 7 subjects. The dataset is split into two sets: a training set and a test set. The training set contains data from five of the actors (S1, S5, S6, S7, S8), while the test set contains data from the remaining two actors (S9 and S11). These training and test sets are balanced, meaning that they contain an equal number of samples for each activity and for each subject. For data preprocessing~\cite{martinez2017simple, zou2020high, zhao2019semantic, xu2021graph}, we apply standard normalization to the 2D and 3D poses before feeding the data to the model. To achieve zero-centering, the hip joint is adopted as the root joint of the 3D poses.

\begin{figure}[!htb]
\centering
\includegraphics[scale=.29]{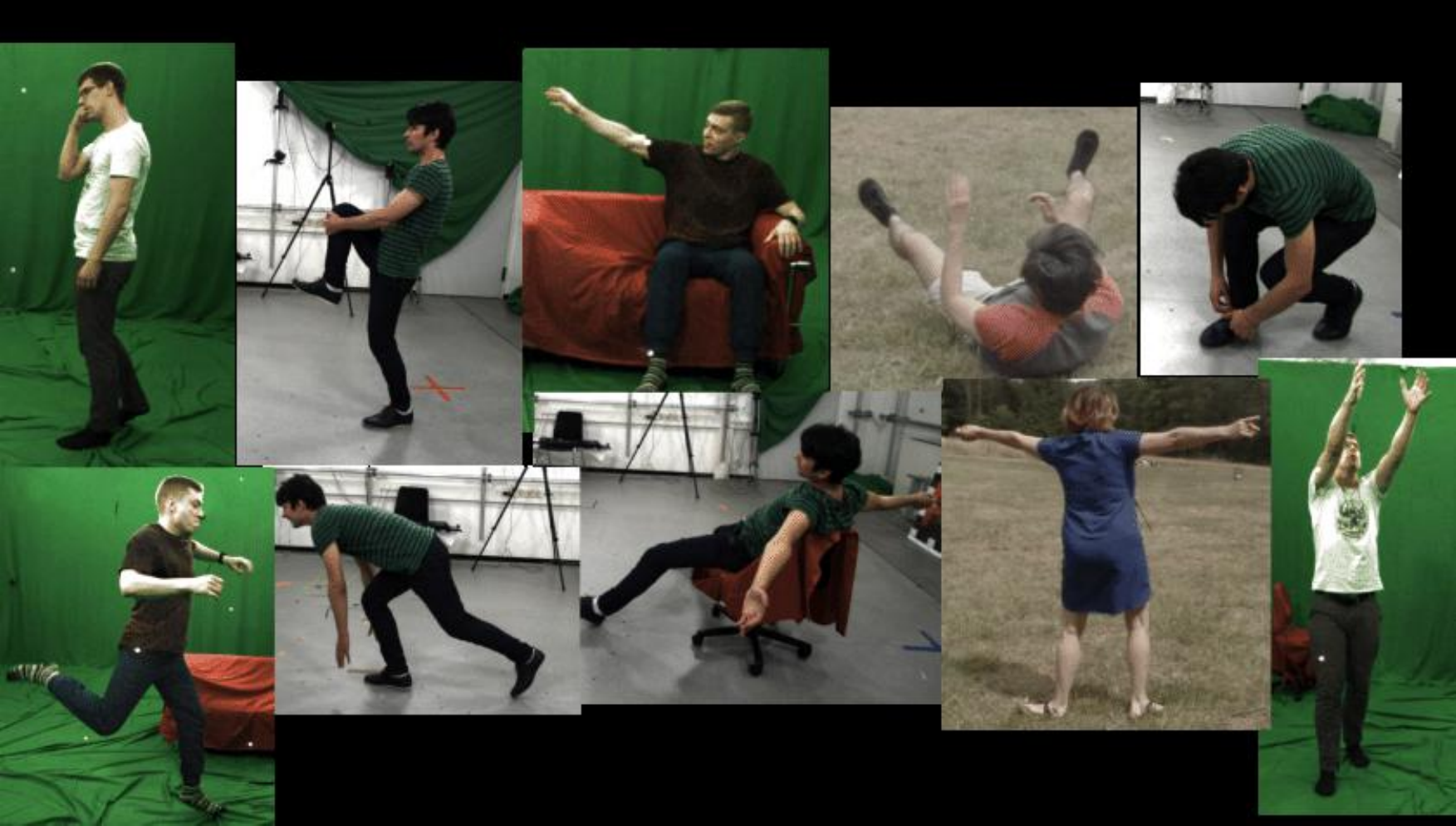}
\caption{Examples of activities in the MPI-INF-3DHP dataset~\cite{mehta2017monocular}.}
\label{Fig:Activities}
\end{figure}

\smallskip
\noindent\textbf{\textsf{\footnotesize MPI-INF-3DHP}} is a benchmark dataset for 3D human pose estimation from monocular RGB images, and comprises both indoor environments with limited space and complex outdoor scenes, as illustrated in Figure~\ref{Fig:Activities}. A total of 8 actors (4 men and 4 women) were captured on camera from 14 camera views, each performing 8 sets of activities that encompassed a wider range of pose categories than the Human3.6M dataset. The activities varied from simple movements such as walking and sitting to more challenging exercises and dynamic actions. The duration of each activity set is approximately one minute, and the actors were dressed in two different sets of clothing that were rotated across the activity sets. One clothing set consisted of casual wear suitable for everyday use, while the other set was plain-colored to facilitate easy augmentation. The dataset also includes ground-truth annotations of the 3D joint positions.

\medskip\noindent\textbf{Evaluation Protocols and Metrics.}\quad For the Human 3.6M benchmark, there are two standard evaluation protocols used for training and testing~\cite{martinez2017simple}, referred to as Protocol \#1 and Protocol \#2. Under Protocol \#1, we report mean per-joint position error (MPJPE), which computes the average Euclidean distance between the predicted and ground-truth 3D positions of each joint after aligning the root joint (i.e., hip joint). Another commonly used metric for evaluating the accuracy of 3D human pose estimation models is the Procrustes-aligned mean per-joint position error (PA-MPJPE), where MPJPE is computed after rigid alignment of the prediction with respect to the ground truth. The PA-MPJPE metric first applies Procrustes analysis to align the predicted and ground-truth joint positions to a common coordinate system. This alignment is performed by scaling, rotating, and translating the predicted joint positions to minimize the sum of squared distances between the predicted and ground-truth joint positions. Once the joint positions are aligned, the PA-MPJPE is calculated by computing the mean of the Euclidean distances between the aligned predicted and ground-truth joint positions for each joint. Both MPJPE and PA-MPJPE are measured in millimeter (mm), and lower error values imply better performance. Both Protocol \#1 and Protocol \#2 use five subjects (S1, S5, S6, S7 and S8) for training and two subjects (S9 and S11) for testing. All camera views are trained with a single model for all actions. For the MPI-INF-3DHP dataset, we use Percentage of Correct Keypoint (PCK) with a threshold of 150mm and Area Under Curve (AUC) for a range of PCK thresholds as evaluation metrics~\cite{habibie2019wild, pavlakos2018ordinal, yang20183d, ci2019optimizing, zeng2021learning, zhu2020deformable}. Both the PCK and AUC metrics provide a measure of how well the predicted joint positions align with the ground-truth joint positions within a certain distance threshold. Higher PCK and AUC scores indicate better performance.

\medskip\noindent\textbf{Baseline Methods.}\quad We compare the performance of our model against various state-of-the-art GCN-based approaches for estimating 3D poses, including SemGCN~\cite{zhao2019semantic}, a GCN-based model that uses a multi-task learning approach to jointly optimize for 3D joint positions and body joint angles; CompGCN~\cite{zou2021compositional}, a hierarchical composition approach that uses a multi-level attention mechanism to adaptively weight the contributions of different body parts at different levels of the hierarchy; High-order GCN~\cite{zou2020high}, a baseline method that employs high-order GCNs to model the complex interactions between body joints; Weight Unsharing~\cite{liu2020comprehensive}, a comprehensive study that analyzes the trade-offs between sharing weights across different body parts versus having separate weights for each body part in GCNs; MM-GCN~\cite{lee2022multi}, a multi-hop GCN-based method that incorporates modulated attention mechanisms to capture the interactions between different body joints across multiple hops; GroupGCN~\cite{zhang2022group}, a decoupling GCN for 3D human pose estimation consisting of group convolution and group interaction; and Modulated GCN~\cite{zou2021modulated}, a GCN-based architecture that employs weight and adjacency modulation mechanisms to capture the complex relationships between different body parts in the human body.

\medskip\noindent\textbf{Implementation Details.}\quad  We implement our model in PyTorch, and conduct all experiments on a single NVIDIA GeForce RTX 3070 GPU with 8GB of memory. For both 2D ground truth and 2D pose detections~\cite{chen2018cascaded}, we train our model for 30 epochs using the AMSGrad optimizer, and we set the initial learning rate to 0.005, the decay factor to 0.65 per 4 epochs, the batch size to 512, and the number of channels to 384. We set the hyperparameter $\beta$ to 0.2, which was computed via grid search with cross-validation on the training set. We also set the weighting factor $\alpha$ to 0.01. We apply dropout with a factor of 0.2 after each graph convolution layer to prevent overfitting. We also integrate a non-local layer~\cite{wang2018non} and a pose refinement module~\cite{cai2019exploiting}. Incorporating an additional pose refinement network, comprised of two fully-connected layers, helps improve performance. However, in the ablation study we exclude the pose refinement network and the non-local layer to ensure a fair comparison.

\subsection{Results and Analysis}
\noindent\textbf{Quantitative Results on Human3.6M.}\quad In Tables~\ref{Tab:MPJPE_Result} and ~\ref{Tab:PA_MPJPE_Result}, we summarize the performance comparison results between our GS-Net model and various state-of-the-art methods for 3D pose estimation on Human3.6M. In both tables, we report results for all 15 actions, as well as the average performance. As can be seen in Table~\ref{Tab:MPJPE_Result}, our method achieves on average 47.1mm and 38.7mm in terms of MPJPE and PA-MPJPE, respectively, outperforming all the baselines. Under Protocol \#1, Table~\ref{Tab:MPJPE_Result} reveals that our GS-Net model performs better than Modulated GCN~\cite{zou2021modulated} in 13 out of 15 actions, yielding 2.3mm error reduction on average, improving upon this best performing baseline by a relative improvement of 4.65\%, while maintaining a fairly small number of learnable parameters. Our method also consistently performs better in almost all actions and outperforms SemGCN~\cite{zhao2019semantic} by a significant relative improvement of 18.23\% on average. Moreover, our model achieves better predictions than the best baseline on challenging actions (i.e., hard poses) that involve self-occlusion such as ``Eating'', ``Sitting'' and ``Smoking'', yielding relative error reductions of 5.9\%, 5.22\% and 7.04\%, respectively, in terms of MPJPE. These self-occlusions can make human pose estimation more challenging because they limit the visible information available to the model. For instance, when eating or smoking, the hands and arms of a person can occlude parts of their face and upper body. Also, when a person is sitting, their legs and arms can occlude other body parts, such as the torso or feet.

\begin{table*}[!htb]
\caption{Performance comparison of our model and baseline methods using MPJPE (in millimeters) between the ground truth and estimated pose on Human3.6M under Protocol \#1. The last column report the average errors. Boldface numbers indicate the best 3D pose estimation performance, whereas the underlined numbers indicate the second best performance.}
\footnotesize
\setlength{\tabcolsep}{3pt}
\smallskip
\centering
\begin{tabular}{l*{17}{c}}
\toprule[1pt]
& \multicolumn{15}{c}{Action}\\
\cmidrule(lr){2-16}
Method & Dire. & Disc. &  Eat & Greet & Phone & Photo &  Pose & Purch. & Sit & SitD. & Smoke & Wait & WalkD. & Walk & WalkT. & Avg.\\
\midrule[.8pt]
Martinez \textit{et al.}~\cite{martinez2017simple} &  51.8 &56.2 &58.1 &59.0 &69.5& 78.4 &55.2 &58.1 &74.0 &94.6 &62.3 &59.1& 65.1& 49.5& 52.4 &62.9\\
Sun \textit{et al.}~\cite{sun2017compositional} &52.8& 54.8& 54.2& 54.3 &61.8 &67.2& 53.1& 53.6 &71.7 &86.7 &61.5 &53.4& 61.6 &47.1& 53.4 &59.1\\
Yang \textit{et al.}~\cite{yang20183d} & 51.5& 58.9& 50.4 &57.0& 62.1& 65.4 &49.8& 52.7& 69.2& 85.2& 57.4& 58.4& 43.6& 60.1& 47.7& 58.6\\
Fang \textit{et al.}~\cite{fang2018learning}& 50.1 &54.3& 57.0& 57.1& 66.6& 73.3& 53.4& 55.7& 72.8& 88.6& 60.3 &57.7& 62.7& 47.5 &50.6& 60.4\\
Hossain \& Little~\cite{hossain2018exploiting}  & 48.4 & 50.7 & 57.2 & 55.2 & 63.1 & 72.6 & 53.0 & 51.7 & 66.1 & 80.9 & 59.0 & 57.3 & 62.4 & 46.6 & 49.6 & 58.3\\
Pavlakos \textit{et al.}~\cite{pavlakos2018ordinal} & 48.5& 54.4& 54.4& 52.0 &59.4 &65.3 &49.9& 52.9& 65.8 &71.1& 56.6& 52.9& 60.9& 44.7& 47.8& 56.2\\
Sharma \textit{et al.}~\cite{sharma2019monocular} & 48.6 &54.5& 54.2& 55.7& 62.2& 72.0& 50.5& 54.3& 70.0& 78.3 &58.1& 55.4& 61.4& 45.2& 49.7& 58.0\\
Zhao \textit{et al.}~\cite{zhao2019semantic} & 47.3& 60.7& 51.4 &60.5& 61.1& \textbf{49.9} & 47.3 & 68.1 &86.2& \textbf{55.0}& 67.8& 61.0& 42.1 & 60.6& 45.3& 57.6\\
Li \textit{et al.}~\cite{li2020weakly} & 62.0 & 69.7 & 64.3 & 73.6 & 75.1 & 84.8 & 68.7 & 75.0 & 81.2 & 104.3 & 70.2 & 72.0 & 75.0 & 67.0 & 69.0 & 73.9\\
Banik \textit{et al.}~\cite{banik20213d} & 51.0 & 55.3 & 54.0 & 54.6 & 62.4 & 76.0 & 51.6 & 52.7 & 79.3 & 87.1 & 58.4 & 56.0 & 61.8 & 48.1 & 44.1 & 59.5\\
Xu \textit{et al.}~\cite{xu2021monocular} & 47.1 & 52.8 & 54.2 & 54.9 & 63.8 & 72.5 & 51.7 & 54.3 & 70.9 & 85.0 & 58.7 & 54.9 & 59.7 & 43.8 & 47.1 & 58.1\\
Zou \textit{et al.}~\cite{zou2020high} & 49.0& 54.5& 52.3& 53.6& 59.2 &71.6& 49.6& 49.8 &66.0 &75.5 &55.1 &53.8& 58.5& 40.9 & 45.4 &55.6\\
Quan \textit{et al.}~\cite{quan2021higher} & 47.0 & 53.7 & 50.9 & 52.4 & 57.8 & 71.3 & 50.2 & 49.1 & 63.5 & 76.3 & 54.1 & 51.6 & 56.5 & 41.7 & 45.3 & 54.8\\
Zou \textit{et al.}~\cite{zou2021compositional} & 48.4 & 53.6 & 49.6 & 53.6 & 57.3 & 70.6 & 51.8 & 50.7 & 62.8 & 74.1 & 54.1 & 52.6 & 58.2 & 41.5 & 45.0 & 54.9\\
Liu \textit{et al.}~\cite{liu2020comprehensive} & 46.3 & 52.2 & 47.3 & 50.7 & 55.5 & 67.1 & 49.2 & 46.0 & 60.4 & 71.1 & 51.5 & 50.1 & 54.5 & 40.3 & 43.7 & 52.4\\
Zou \textit{et al.}~\cite{zou2021modulated} & 45.4 & \underline{49.2} & \underline{45.7} & \underline{49.4} & \underline{50.4} & 58.2 & \underline{47.9} & \underline{46.0} & \underline{57.5} & \underline{63.0} & \underline{49.7} & \underline{46.6} & \underline{52.2} & \underline{38.9} & \textbf{40.8} & \underline{49.4}\\
Lee \textit{et al.}~\cite{lee2022multi} & 46.8 & 51.4 & 46.7 & 51.4 & 52.5 & 59.7 & 50.4 & 48.1 & 58.0 & 67.7 & 51.5 & 48.6 & 54.9 & 40.5 & 42.2 & 51.7\\
Zhang \textit{et al.}~\cite{zhang2022group} & \underline{45.0} & 50.9 & 49.0 & 49.8 & 52.2 & 60.9 & 49.1 & 46.8 & 61.2 & 70.2 & 51.8 & 48.6 & 54.6 & 39.6 & 41.2 & 51.6\\
\midrule[.8pt]
Ours & \textbf{41.1} & \textbf{46.6} & \textbf{43.0} & \textbf{48.0} &\textbf{48.6} & \underline{52.4} & \textbf{44.6} & \textbf{41.9} & \textbf{54.5} & 65.9  &\textbf{46.2} & \textbf{46.1} & \textbf{48.2} & \textbf{38.6} & \underline{40.9} & \textbf{47.1} \\
%Ours (GT) & 32.7 & 41.0 & 32.1 & 36.9 & 37.6 & 41.7 & 39.4 & 35.9 & 42.8 & 44.9 & 36.9 & 37.8 & 38.4 & 30.5 & 32.7 & 37.4\\
\bottomrule[1pt]
\end{tabular}
\label{Tab:MPJPE_Result}
\end{table*}

Under Protocol \#2, Table~\ref{Tab:PA_MPJPE_Result} shows that our model on average reduces the error by 1.83\% compared to Modulated GCN~\cite{zou2021modulated}, and achieves better results in 12 out of 15 actions. Also, our method outperforms Modulated GCN on the challenging actions of ``Eating'', ``Sitting'' and ``Smoking'', yielding relative error reductions of 3.58\%, 4.74\% and 5.68\%, respectively, in terms of PA-MPJPE. Moreover, our model performs better than Modulated GCN on the challenging ``Photo'' action, yielding a relative error reduction of 4.72\%. In addition, GS-Net outperforms High-order GCN~\cite{zou2020high} by a relative improvement of 4.86\% on average, as well as on all actions.

\begin{table*}[!htb]
\caption{Performance comparison of our model and baseline methods using PA-MPJPE between the ground truth and estimated pose on Human3.6M under Protocol \#2.}
\footnotesize
\setlength{\tabcolsep}{3pt}
\smallskip
\centering
\begin{tabular}{l*{17}{c}}
\toprule[1pt]
& \multicolumn{15}{c}{Action}\\
\cmidrule(lr){2-16}
Method & Dire. & Disc. &  Eat & Greet & Phone & Photo &  Pose & Purch. & Sit & SitD. & Smoke & Wait & WalkD. & Walk & WalkT. & Avg.\\
\midrule[.8pt]
Pavlakos \textit{et al.}~\cite{pavlakos2017coarse} & 47.5 &50.5 &48.3& 49.3& 50.7 &55.2 &46.1 &48.0& 61.1& 78.1 &51.1& 48.3& 52.9& 41.5& 46.4 &51.9 \\
Zhou \textit{et al.}~\cite{zhou2017towards} & 47.9& 48.8 &52.7& 55.0& 56.8& 49.0 &45.5 &60.8& 81.1 &\underline{53.7}& 65.5& 51.6& 50.4 &54.8 &55.9& 55.3\\
Martinez \textit{et al.}~\cite{martinez2017simple} & 39.5 &43.2 &46.4 &47.0 &51.0& 56.0 &41.4& 40.6 &56.5& 69.4& 49.2& 45.0& 49.5& 38.0 &43.1 &47.7\\
Sun \textit{et al.}~\cite{sun2017compositional} & 42.1& 44.3& 45.0 &45.4 &51.5 &53.0 &43.2& 41.3& 59.3 &73.3& 51.0& 44.0& 48.0& 38.3& 44.8& 48.3\\
Fang \textit{et al.}~\cite{fang2018learning} & 38.2& 41.7& 43.7& 44.9& 48.5 &55.3& 40.2& 38.2& 54.5 &64.4& 47.2 &44.3& 47.3& 36.7& 41.7& 45.7\\
Hossain \& Little~\cite{hossain2018exploiting}  & 35.7 & 39.3 & 44.6 &43.0& 47.2& 54.0& 38.3 &37.5 &51.6 &61.3& 46.5& 41.4 &47.3 &34.2 &39.4& 44.1\\
Lee \textit{et al.}~\cite{lee2018propagating}  & 38.0 & 39.3 & 46.3 & 44.4 & 49.0 & 55.1 & 40.2 & 41.1 & 53.2 & 68.9 & 51.0 & 39.1 & \textbf{33.9} & 56.4 & 38.5 & 46.2 \\
Li \textit{et al.}~\cite{li2020weakly} & 38.5 & 41.7 & 39.6 & 45.2 & 45.8 & 46.5 & 37.8 & 42.7 & 52.4 & 62.9 & 45.3 & 40.9 & 45.3 & 38.6 & 38.4 & 44.3\\
Banik \textit{et al.}~\cite{banik20213d} & 38.4 & 43.1 & 42.9 & 44.0 & 47.8 & 56.0 & 39.3 & 39.8 & 61.8 & 67.1 & 46.1 & 43.4 & 48.4 & 40.7 & 35.1 & 46.4\\
Xu \textit{et al.}~\cite{xu2021monocular} & 36.7 & 39.5 & 41.5 & 42.6 & 46.9 & 53.5 & 38.2 & 36.5 & 52.1 & 61.5 & 45.0 & 42.7 & 45.2 & 35.3 & 40.2 & 43.8\\
Zou \textit{et al.}~\cite{zou2020high} &38.6 &42.8& 41.8 &43.4 &44.6& 52.9& \textbf{37.5}& 38.6 &53.3 &60.0& 44.4& 40.9& 46.9 &32.2 &37.9 &43.7\\
Quan \textit{et al.}~\cite{quan2021higher} & 36.9 & 42.1 & 40.3 & 42.1 & 43.7 & 52.7 & 37.9 & 37.7 & 51.5 & 60.3 & 43.9 & 39.4 & 45.4 & 31.9 & 37.8 & 42.9\\
Zou \textit{et al.}~\cite{zou2021compositional} & 38.4 & 41.1 & 40.6 & 42.8 & 43.5 & 51.6 & 39.5 & 37.6 & 49.7 & 58.1 & 43.2 & 39.2 & 45.2 & 32.8 & 38.1 & 42.8\\
Liu \textit{et al.}~\cite{liu2020comprehensive} & 35.9 & 40.0 & 38.0 & 41.5 & 42.5 & 51.4 &  37.8 & 36.0 & 48.6 & 56.6 & 41.8 & 38.3 & 42.7 & 31.7 & 36.2 & 41.2\\
Zou \textit{et al.}~\cite{zou2021modulated} & 35.7 & \underline{38.6} & \underline{36.3} & \textbf{40.5} & \underline{39.2} & \underline{44.5} & 37.0 & 35.4 & \underline{46.4} & \textbf{51.2} & \underline{40.5} & \textbf{35.6} & 41.7 & \underline{30.7} & 33.9 & \underline{39.1}\\
Lee \textit{et al.}~\cite{lee2022multi} & 35.7 & 39.6 & 37.3 & 41.4 & 40.0 & 44.9 & 37.6 & 36.1 & 46.5 & 54.1 & 40.9 & 36.4 & 42.8 & 31.7 & 34.7 & 40.3\\
Zhang \textit{et al.}~\cite{zhang2022group} & \underline{35.3} & 39.3 & 38.4 & 40.8 & 41.4 & 45.7 & 36.9 & \underline{35.1} & 48.9 & 55.2 & 41.2 & 36.3 & 42.6 & 30.9 & \underline{33.7} & 40.1\\
\midrule[.8pt]
Ours & \textbf{34.5} & \textbf{38.4} &\textbf{35.0} & \underline{40.9} & \textbf{38.9} & \textbf{42.4} & \underline{35.9} & \textbf{33.9} & \textbf{44.2} & 55.9  &\textbf{38.2} & \underline{36.7} & \underline{40.6} & \textbf{30.4} & \textbf{33.8} & \textbf{38.7} \\
%Ours (GT) & 24.8 & 30.5 & 25.7 & 29.0 & 28.4 & 32.0 & 30.4 & 26.0 & 32.36 & 36.0 & 29.4 & 28.4 & 30.7 & 24.1 & 26.3 & 29.0\\
\bottomrule[1pt]
\end{tabular}
\label{Tab:PA_MPJPE_Result}
\end{table*}

\medskip\noindent\textbf{Cross-Dataset Results on MPI-INF-3DHP.}\quad In Table~\ref{Tab:mpi_3dhp_inf}, we compare our method against strong baselines to test its generalization ability across different datasets. We train our model on the Human3.6M dataset and test it on the MPI-INF-3DHP dataset. The results show that our approach achieves the highest PCK and AUC scores, consistently yielding superior performance over the baseline methods in both indoor and outdoor scenes. Compared to the best performing baseline, our model yields relative improvements of 2.68\% and 5.37\% in terms of the PCK and AUC metrics, respectively. Although our model is only trained with indoor scenes on Human3.6M, it produces satisfactory results with outdoor settings. This verifies the strong generalization ability of our approach to unseen scenarios and datasets.

\begin{table}[!htb]
\caption{Performance comparison of our model and baseline methods on the MPI-INF-3DHP dataset using PCK and AUC as evaluation metrics. Higher values in boldface indicate the best performance, whereas the underlined numbers indicate the second best performance. }
\small
\setlength{\tabcolsep}{3pt}
\smallskip
\centering
\begin{tabular}{lcc}
\toprule[1pt]
Method & PCK $(\uparrow)$ & AUC $(\uparrow)$\\
\midrule[.8pt]
Martinez \textit{et al.}~\cite{martinez2017simple} & 42.5 & 17.0 \\
Mehta \textit{et al.}~\cite{mehta2017monocular} & 64.7 & 31.7 \\
Li \textit{et al.}~\cite{li2019generating} & 67.9 & - \\
Yang \textit{et al.}~\cite{yang20183d} & 69.0 & 32.0 \\
Zhou \textit{et al.}~\cite{zhou2017towards} & 69.2 & 32.5 \\
Habibie \textit{et al.}~\cite{habibie2019wild} & 70.4 & 36.0 \\
Pavlakos \textit{et al.}~\cite{pavlakos2018ordinal} & 71.9 & 35.3 \\
Wang \textit{et al.}~\cite{wang2019not} & 71.9 & 35.8 \\
Quan \textit{et al.}~\cite{quan2021higher} & 72.8 & 36.5 \\
Ci \textit{et al.}~\cite{ci2019optimizing} & 74.0 & 36.7 \\
Zhou \textit{et al.}~\cite{zhou2019hemlets} & 75.3 & 38.0 \\
Zeng \textit{et al.}~\cite{zeng2020srnet} & 77.6 & 43.8 \\
Liu \textit{et al.}~\cite{liu2020comprehensive} & 79.3 & 47.6 \\
Zhou \textit{et al.}~\cite{zou2021compositional} & 79.3 & 45.9 \\
Xu \textit{et al.}~\cite{xu2021graph} & 80.1 & 45.8 \\
Zeng \textit{et al.}~\cite{zeng2021learning} & \underline{82.1} & 46.2 \\
Lee \textit{et al.}~\cite{lee2022multi} & 81.6 & \underline{50.3}\\
Zhang \textit{et al.}~\cite{zhang2022group} & 81.1 & 49.9\\
\midrule[.8pt]
Ours & \textbf{84.3} & \textbf{53.0}\\
\bottomrule[1pt]
\end{tabular}
\label{Tab:mpi_3dhp_inf}
\end{table}

\begin{figure*}[!htb]
\small
\centering
\includegraphics[scale=1]{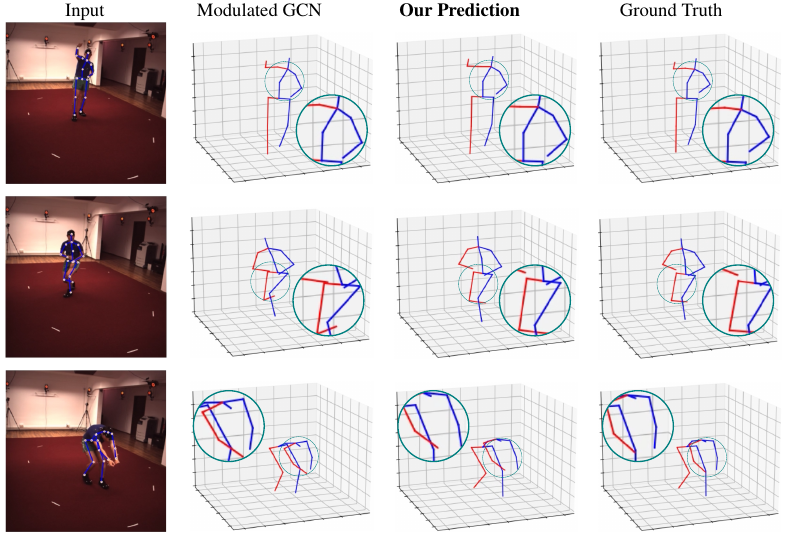}
\includegraphics[scale=.8]{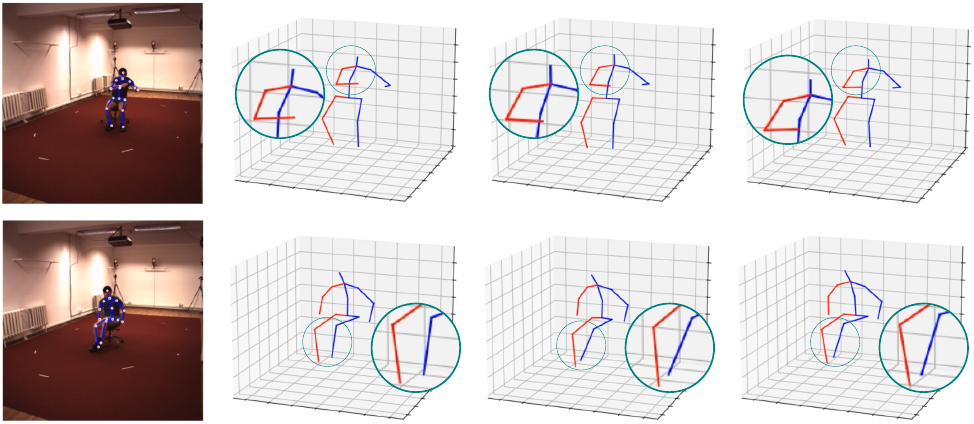}
\caption{Visual comparison between GS-Net, Modulated GCN and ground truth on the Human3.6M test set. Compared to Modulated GCN, our model is able to produce better predictions.}
\label{Fig:Qualitative}
\end{figure*}

\medskip\noindent\textbf{Qualitative Results.}\quad In Figure~\ref{Fig:Qualitative}, we show the visual results obtained by GS-Net on the Human3.6M dataset for various actions. The effectiveness of our proposed approach in addressing the 2D-to-3D pose estimation task is demonstrated by the close match between the inferred 3D poses and the ground truth, as evidenced by the accurate results produced by our model from input images. Compared to Modulated GCN, our model yields pose predictions that are closer to the ground truth, even when dealing with challenging actions that involve self-occlusion.

\medskip\noindent\textbf{Quantitative Results using Ground Truth.}\quad We also compare our model with GCN-based methods, including SemGCN~\cite{zhao2019semantic}, High-order GCN~\cite{zou2020high}, HOIF-Net~\cite{quan2021higher}, Weight Unsharing~\cite{liu2020comprehensive}, and Modulated GCN~\cite{zou2021modulated} using ground truth. The results are reported in Table~\ref{Tab:baselineComparison}, which shows that our model consistently yields better performance than GCN-based approaches under both Protocols \#1 and \#2. Under Protocol \#1, our model outperforms SemGCN, High-order GCN, HOIF-Net, Modulated GCN, and Weight Unsharing by 4.73mm, 2.11mm, 0.71mm, 0.84mm, and 0.42mm, respectively, resulting in relative error reductions of 11.22\%, 5.34\%, 1.86\%, 2.20\%, and 1.11\%. Under Protocol \#2, our model also outperforms SemGCN, High-order GCN, HOIF-Net, Modulated GCN, and Weight Unsharing by 4.59mm, 2.13mm, 0.8mm, 1.12mm, and 1.15mm, with relative error reductions of 13.69\%, 6.86\%, 2.69\%, 3.73\%, and 3.82\%, respectively.

\begin{table}[!htb]
\caption{Performance comparison of our model and other state-of-the-art GCN-based methods. Our proposed GS-Net method achieves the best performance, as indicated by boldface numbers. All errors are measured in millimeters (mm).}
\small
\setlength{\tabcolsep}{3pt}
\smallskip
\centering
\begin{tabular}{lcc}
\toprule[1pt]
Method & MPJPE $(\downarrow)$ & PA-MPJPE $(\downarrow)$ \\
\midrule[.8pt]
SemGCN~\cite{zhao2019semantic} & 42.14 & 33.53 \\
High-order GCN~\cite{zou2020high} & 39.52 & 31.07\\
HOIF-Net~\cite{quan2021higher} & 38.12 & 29.74\\
Modulated GCN~\cite{zou2021modulated} & 38.25 & 30.06\\
Weight Unsharing~\cite{liu2020comprehensive} & 37.83 & 30.09\\
\midrule[.8pt]
Ours  & \textbf{37.41} & \textbf{28.94} \\
\bottomrule[1pt]
\end{tabular}
\label{Tab:baselineComparison}
\end{table}

\subsection{Ablation Study}
In order to better understand the impact of the design choices made for our network architecture, we perform ablation experiments on Human3.6M under controlled settings. We report the results using 2D ground truth for training and testing to eliminate the added uncertainty from 2D pose detectors. These ablation studies demonstrate the efficacy of the key components of our model.

\medskip\noindent\textbf{Impact of Skip Connection.}\quad  We analyze how the weighted initial skip connection in the layer-wise propagation rule affects the model performance, and we report the results in Table~\ref{Tab:skipConnection}. We can see that our model benefits from the weighted initial skip connection, yielding relative error reductions of 3.43\% and 2.91\% in terms of MPJPE and PA-MPJPE, respectively.

\begin{table}[!htb]
\caption{Effectiveness of initial skip connection (ISC). Boldface numbers indicate better performance.}
\small
\setlength{\tabcolsep}{2.5pt}
\smallskip
\centering
\begin{tabular}{l*{7}{c}}
\toprule
Method & MPJPE $(\downarrow)$ & PA-MPJPE $(\downarrow)$ \\
\midrule
w/o ISC & 38.74 & 29.79 \\
w ISC & \textbf{37.41} & \textbf{28.94}\\
\bottomrule

\end{tabular}
\label{Tab:skipConnection}
\end{table}

\medskip\noindent\textbf{Impact of Residual Block Design.}\quad We compare two residual block designs and report the results in the Table~\ref{Tab:residualBlockDesign}. The first design employs blocks consisting of convolutional layers, followed by batch normalization (BatchNorm) and a ReLU activation function. By contrast, the second design uses blocks comprised of convolutional layers, followed by layer normalization (LayerNorm) and a GELU activation function. We incorporate the latter design into the proposed architecture, resulting in improved performance compared to the first design. The results show that our model with ConvNeXt block achieves a 0.63\% decrease in error in terms of the MPJPE metric and yields comparable performance in terms of the PA-MPJPE metric. In this part of the ablation study, we supply 2D keypoints obtained from the 2D pose detector as input to our model to examine the model reliability. We also include the pose refinement module and the non-local layer in the network.

\begin{table}[!htb]
\caption{Effect of residual block design on the performance of our model. Lower values in boldface indicate better performance.}
\small
\setlength{\tabcolsep}{3pt}
\smallskip
\centering
\begin{tabular}{lcc}
\toprule[1pt]
Method & MPJPE $(\downarrow)$ & PA-MPJPE $(\downarrow)$ \\
\midrule[.8pt]
Ours w/ BatchNorm and ReLU & 47.40 & \textbf{38.62} \\
Ours w/ LayerNorm and GELU & \textbf{47.10} & 38.65\\
\bottomrule[1pt]
\end{tabular}
\label{Tab:residualBlockDesign}
\end{table}

\begin{figure}[!htb]
\centering
\begin{tabular}{c}
\includegraphics[scale=.52]{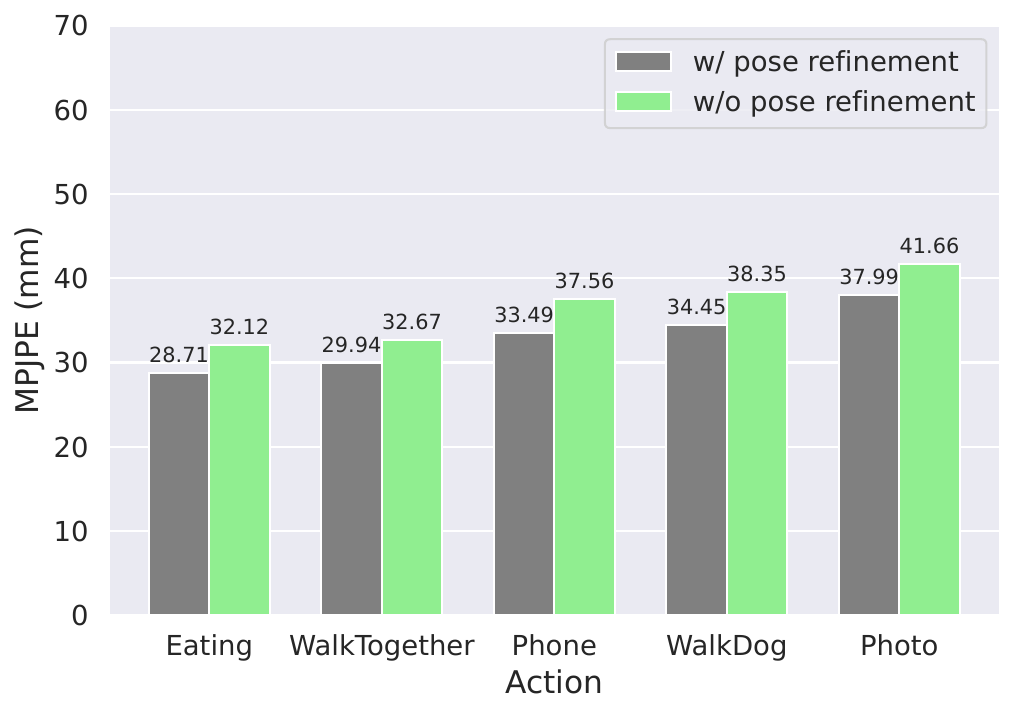} \\
\includegraphics[scale=.52]{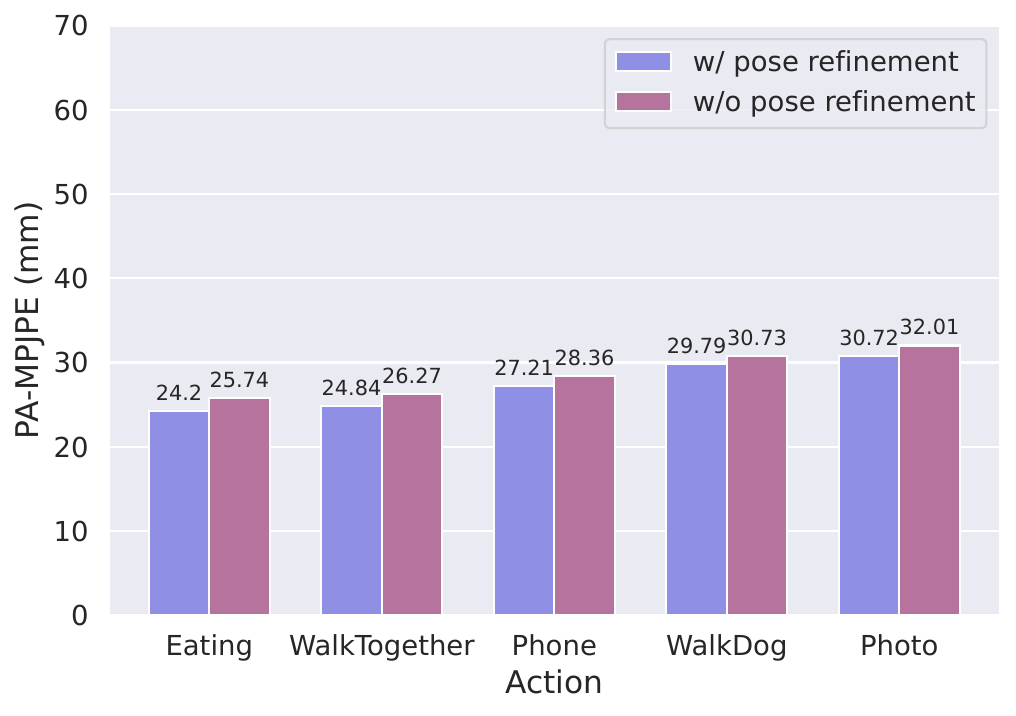}
\end{tabular}
\caption{Performance comparison of our model with and without pose refinement using MPJPE (top) and PA-MPJPE (bottom). When coupled with a pose refinement network, GS-Net performs consistently better on challenging actions.}
\label{Fig:poseRefinement}
\end{figure}

\medskip\noindent\textbf{Impact of Pose Refinement.}\quad We also scrutinize the effectiveness of the pose refinement network. The results in Table~\ref{Tab:poseRefinementTab} show that on average the MPJPE and PA-MPJPE errors are reduced by 3.74mm and 1mm, respectively, demonstrating the advantage of using pose refinement in achieving better performance under both protocols. To reinforce our claim, we report our findings in Figure~\ref{Fig:poseRefinement} that shows the performance comparison with or without the pose refinement model under Protocol \#1 (top) and Protocol \#2 (bottom) for various challenging actions such as ``Eating'' and ``Photo''. For example, relative error reductions of 10.62\% and 5.98\%  are achieved for the ``Eating'' action in terms of MPJPE and PA-MPJPE, respectively.

\begin{table}[!htb]
\caption{Effectiveness of the pose refinement network (PRN). Boldface numbers indicate better performance.}
\small
\smallskip
\centering
\begin{tabular}{lcc}
\toprule[1pt]
Method & MPJPE $(\downarrow)$ & PA-MPJPE $(\downarrow)$ \\
\midrule[.8pt]
w/o PRN & 37.41 & 28.94 \\
w PRN & \textbf{33.67} & \textbf{27.94}\\
\bottomrule[1pt]
\end{tabular}
\label{Tab:poseRefinementTab}
\end{table}

\medskip\noindent\textbf{Impact of Symmetrizing Adjacency Modulation.}\quad As reported in Table~\ref{Tab:symmetryRegularizationTab}, symmetrizing the modulated adjacency modulation matrix helps reduce the MPJPE and PA-MPJPE errors by 1.25mm and 0.48mm, respectively.

\begin{table}[!htb]
\caption{Effectiveness of symmetrizing adjacency modulation. Boldface numbers indicate better performance.}
\small
\smallskip
\centering
\begin{tabular}{l*{7}{c}}
\toprule[1pt]
Method & MPJPE $(\downarrow)$  & PA-MPJPE $(\downarrow)$ \\
\midrule[.8pt]
w/o Symmetry & 38.66 & 29.42 \\
w Symmetry & \textbf{37.41} & \textbf{28.94}\\
\bottomrule[1pt]
\end{tabular}
\label{Tab:symmetryRegularizationTab}
\end{table}

\medskip\noindent\textbf{Impact of Loss Functions.}\quad The results in Table~\ref{Tab:lossFunctionsTab} validate our design decision to adopt a weighted loss function comprised of $\ell_1$ and $\ell_2$ penalty terms. It is evident that using the weighted sum of both penalty terms results in better performance in terms of MPJPE and PA-MPJPE. This better performance is largely attributed to the fact that using a weighted sum of both loss functions helps balance the trade-offs between robustness and sensitivity to smaller errors. In other words, we can take advantage of the robustness of $\ell_1$ loss while still providing some level of sensitivity to large errors.

\begin{table}[!htb]
\caption{Effectiveness of the loss functions. Boldface numbers indicate better performance.}
\small
\setlength{\tabcolsep}{3pt}
\smallskip
\centering
\begin{tabular}{l*{7}{c}}
\toprule[1pt]
Method & MPJPE $(\downarrow)$ & PA-MPJPE $(\downarrow)$ \\
\midrule[.8pt]
Only $\ell_1$ loss & 37.43 & 29.29 \\
Only $\ell_2$ loss & 37.62 & 29.25 \\
Weighted sum of $\ell_1$ and $\ell_2$ losses & \textbf{37.41} & \textbf{28.94}\\
\bottomrule[1pt]
\end{tabular}
\label{Tab:lossFunctionsTab}
\end{table}

\begin{figure}[!htb]
\centering
\setlength{\tabcolsep}{5pt}
\begin{tabular}{c}
\includegraphics[width=3.3in]{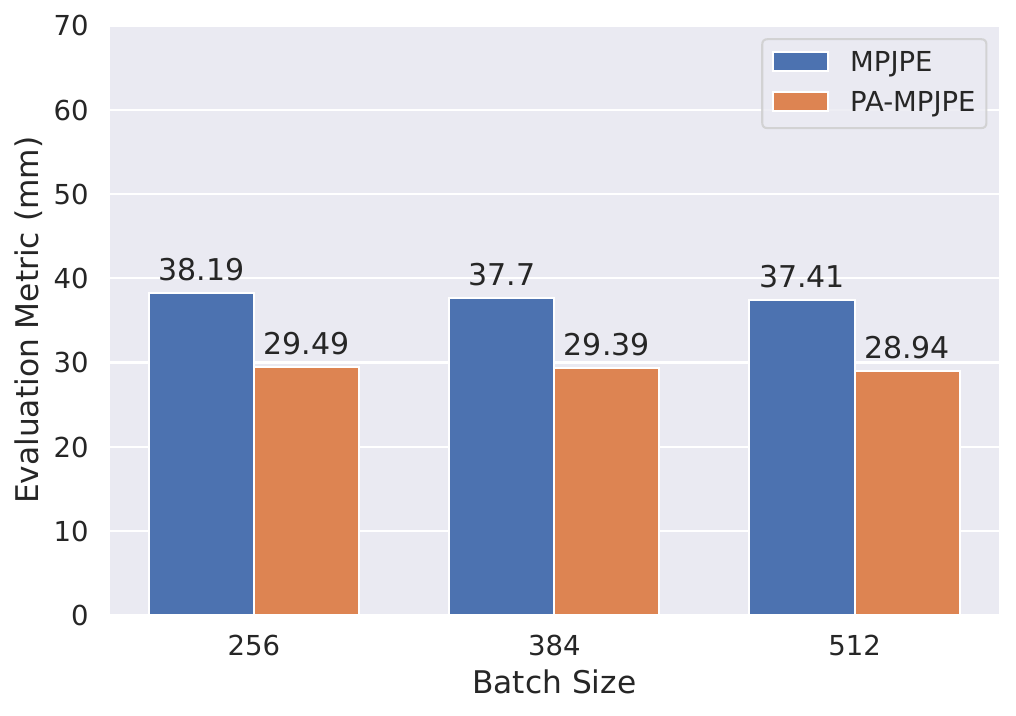} \\
\includegraphics[width=3.3in]{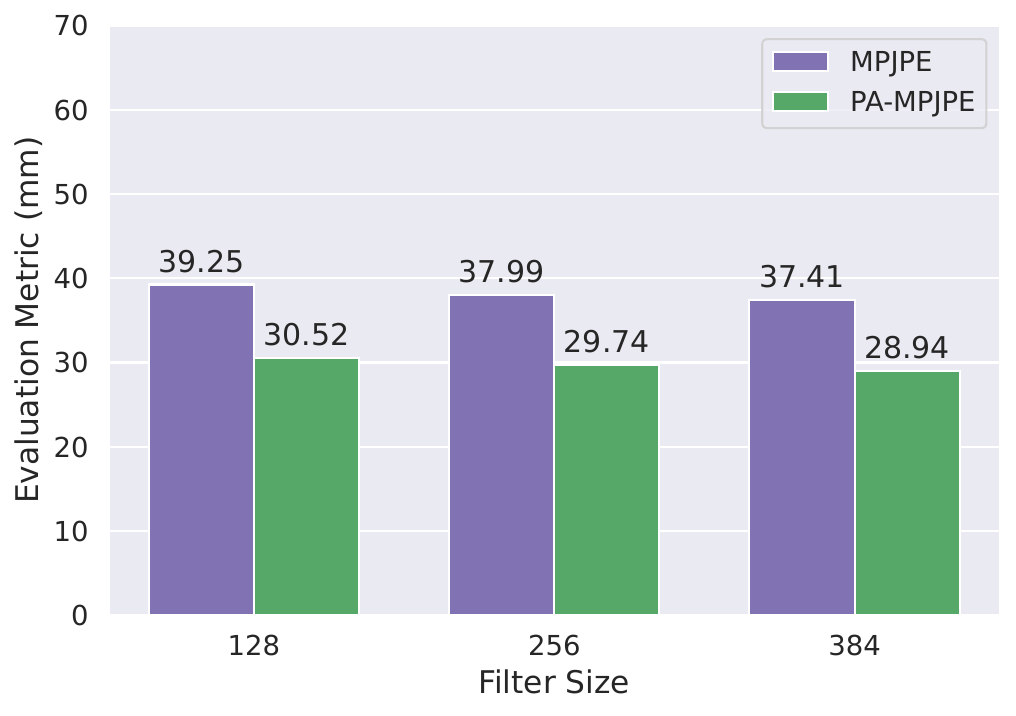}
\end{tabular}
\caption{Performance of our proposed GS-Net model on the Human3.6M dataset using varying batch and filter sizes.}
\label{Fig:FigureBatchFilter}
\end{figure}

\medskip\noindent\textbf{Impact of Batch/Filter Size.}\quad In Figure~\ref{Fig:FigureBatchFilter}, we analyze the effect of varying batch and filter sizes on our model performance. We can see that a batch size of 512 and a filter size of 384 yield the best results in terms of MPJPE and PA-MPJPE, respectively.

\begin{figure}[!htb]
\centering
\setlength{\tabcolsep}{5pt}
\begin{tabular}{cc}
\includegraphics[scale=.51]{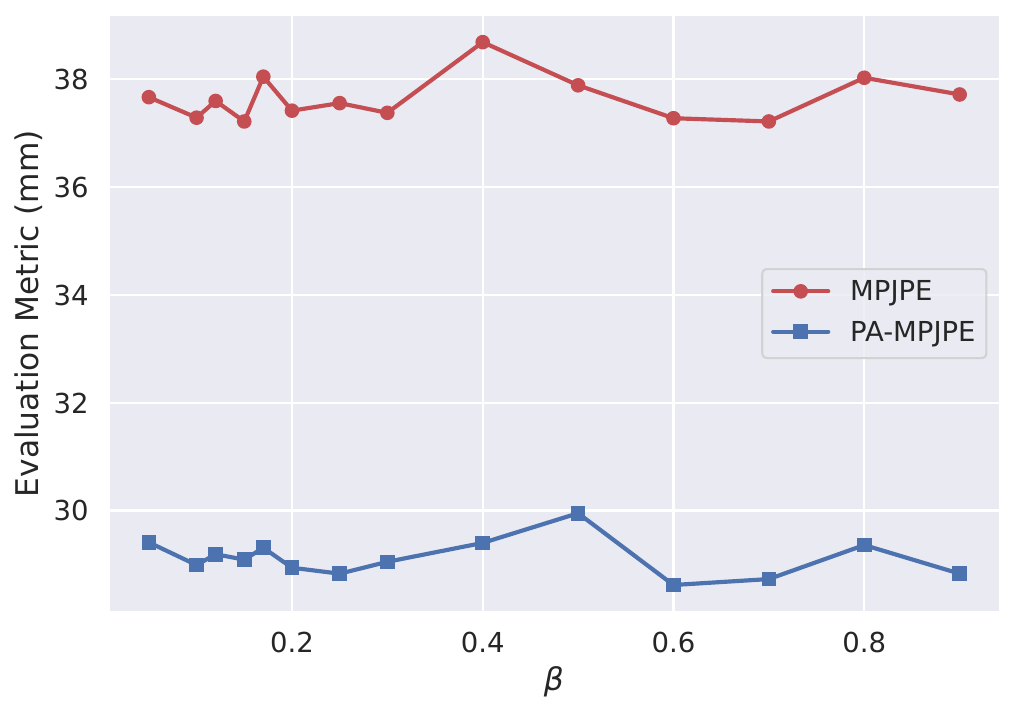}
\end{tabular}
\caption{Sensitivity analysis of our model to the choice of the Laplacian regularization hyperparameter $\beta$. Smaller values of $\beta$ generally result in lower MPJPE and PA-MPJPE errors.}
\label{Fig:FigureHyperparameter}
\end{figure}

\medskip\noindent\textbf{Hyperparameter Sensitivity Analysis.}\quad We also examine the influence of the Laplacian regularization hyperparameter $\beta$ on model performance by plotting the error metrics vs. $\beta$ for a range of values in the interval $(0,1)$. This hyperparameter controls the strength of the Laplacian regularization term in the objective function. It determines how much emphasis is placed on the regularization term relative to the data fidelity term in the objective function. A larger value of the regularization parameter corresponds to a stronger regularization effect, places more emphasis on smoothness in the filtered feature vectors of neighboring graph nodes, but at the expense of how well the filtered feature vectors match the initial feature vectors. Conversely, a smaller value of $\beta$ places more emphasis on the data fidelity term and results in a smaller error between the initial and filtered feature vectors. As shown in Figure~\ref{Fig:FigureHyperparameter}, the best results are typically obtained when the regularization parameter is small. We can observe that our model achieves the lowest error values of MPJPE and PA-MPJPE when $\beta = 0.2$, respectively.

\subsection{Runtime Analysis}
We also analyze the model's inference time, which is a crucial factor in determining the efficiency of the performance of our proposed approach. By examining the inference time, we aim to understand the speed at which our model processes and generates the output. The inference time results are reported in Table~\ref{Tab:Runtime}, which shows that our model significantly outperforms strong baselines.

\begin{table}[!htb]
\caption{Runtime analysis of our model in comparison with competing baselines.}
\small
\setlength{\tabcolsep}{2.2pt}
\smallskip
\centering
\begin{tabular}{lc}
\toprule[1pt]
Method &  Inference Time \\
\midrule[.8pt]
SemGCN~\cite{zhao2019semantic} & .012s \\
High-Order GCN~\cite{zou2020high}  & .013s \\
HOIF-Net~\cite{quan2021higher} & .016s\\
Weight Unsharing~\cite{liu2020comprehensive} & .032s \\
MM-GCN~\cite{lee2022multi} & .009s\\
Modulated GCN~\cite{zou2021modulated} & .010s \\
\midrule[.8pt]
Ours & .003s \\
\bottomrule[1pt]
\end{tabular}
\label{Tab:Runtime}
\end{table}

\subsection{Limitations}
\begin{figure}[!htb]
\centering
\setlength{\tabcolsep}{5pt}
\includegraphics[scale=.51]{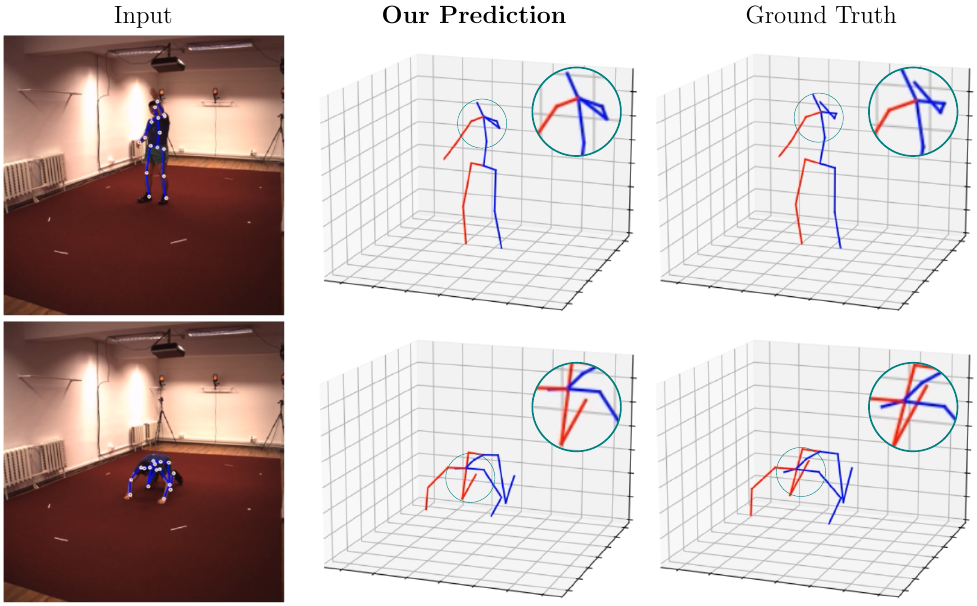}
\caption{Example of the failure cases of our model on the ``Greeting'' and ``Sitting Down'' actions from Human3.6M.}
\label{Fig:FailureCases}
\end{figure}

In addition to the qualitative results depicted in Figure~\ref{Fig:Qualitative}, we thoroughly investigate several instances where our GS-Net model did not perform as expected. We scrutinize these failure cases to gain a deeper understanding of the limitations of our model. Figure~\ref{Fig:FailureCases} illustrates failure cases of our model predictions for the ``Greeting'' and ``Sitting Down'' actions from Human3.6M. As can be seen, our predictions do not align perfectly with the ground truth poses in situations where self-occlusions occur. It is important to point out that these failure cases are not specific to GS-Net alone but rather a common challenge encountered in previous works as well~\cite{zhao2019semantic, zou2021modulated}. This is attributed in part to the diverse actions performed in different ways within the Human3.6M training dataset. In addition, since our model does not incorporate temporal information, the uncertainty inherent in human motion further adds complexity to the prediction process.

\section{Conclusion}
In this paper, we presented a simple yet effective Gauss-Seidel graph neural network together with weight and adjacency modulation. The layer-wise propagation rule of our proposed framework is inspired by the iterative solution of graph filtering with Laplacian regularization using the Gauss-Seidel method. Our network architecture leverages the ConvNeXt residual block, which makes the network more computationally efficient and reduces the risk of overfitting, where the model learns to fit the training data too closely. Empirical experiments show that our model achieves state-of-the-art performance on two benchmark datasets, and can serve as a strong baseline for 3D human pose estimation. We also conducted extensive ablation studies to analyze the impact of different design choices on the model performance. For future work, we plan to explore the applicability of our model to other computer vision and graph representation learning tasks, as well as to improve its computational efficiency and interpretability.

\bibliographystyle{ieeetr}
\bibliography{references}
\end{document}